\documentclass[twoside,11pt]{article}
\usepackage{arxiv}

\usepackage{amsmath,amsfonts,amssymb,amsopn,amsbsy}
\usepackage{dsfont,bm,bbm,times,url,verbatim,epstopdf,xspace}
\usepackage{placeins}
\usepackage[capitalize]{cleveref}
\usepackage[group-separator={,}]{siunitx}
\usepackage[export]{adjustbox}
\usepackage{hhline}
\usepackage[hypertexnames=false]{hyperref}

\newcommand{\ie}{i.e.}
\newcommand{\eg}{e.g.}
\newcommand{\etal}{et al.}
\def\ddefloop#1{\ifx\ddefloop#1\else\ddef{#1}\expandafter\ddefloop\fi}

\def\ddef#1{\expandafter\def\csname bb#1\endcsname{\ensuremath{\mathbb{#1}}}}
\ddefloop ABCDEFGHIJKLMNOPQRSTUVWXYZ\ddefloop

\def\ddef#1{\expandafter\def\csname bf#1\endcsname{\ensuremath{\mathbf{#1}}}}
\ddefloop ABCDEFGHIJKLMNOPQRSTUVWXYZabcdefghijklmnopqrstuvwxyz\ddefloop

\def\ddef#1{\expandafter\def\csname bf#1\endcsname{\ensuremath{\pmb{\csname #1\endcsname}}}}
\ddefloop {alpha}{beta}{gamma}{delta}{epsilon}{varepsilon}{zeta}{eta}{theta}{vartheta}{iota}{kappa}{lambda}{mu}{nu}{xi}{pi}{varpi}{rho}{varrho}{sigma}{varsigma}{tau}{upsilon}{phi}{varphi}{chi}{psi}{omega}{Gamma}{Delta}{Theta}{Lambda}{Xi}{Pi}{Sigma}{varSigma}{Upsilon}{Phi}{Psi}{Omega}{ell}\ddefloop

\def\ddef#1{\expandafter\def\csname c#1\endcsname{\ensuremath{\mathcal{#1}}}}
\ddefloop ABCDEFGHIJKLMNOPQRSTUVWXYZ\ddefloop

\DeclareMathOperator*{\argmax}{arg\,max}

\newcommand\Parens[1]{\left(#1\right)}

\newcommand\Dotp[1]{\left\langle#1\right\rangle}

\newcommand{\RR}{\ensuremath{\bbR}} 

\newcommand{\kernel}{\ensuremath{K}} 

\usepackage[textsize=tiny]{todonotes}

\usepackage{tabularx}
\newcolumntype{Y}{>{\centering\arraybackslash}X}
\usepackage{multirow}

\usepackage{algorithm}
\usepackage{algorithmic}
\usepackage{pdflscape}

\newcommand{\vct}[1]{\boldsymbol{#1}} 

\newcommand{\field}[1]{\mathbb{#1}}
\newcommand{\R}{\field{R}} 
\newcommand{\T}{^{\textrm T}} 

\newcommand{\ProbOpr}[1]{\mathbb{#1}}

\newcommand{\expect}[2]{%
\ifthenelse{\equal{#2}{}}{\ProbOpr{E}_{#1}}
{\ifthenelse{\equal{#1}{}}{\ProbOpr{E}\left[#2\right]}{\ProbOpr{E}_{#1}\left[#2\right]}}} 
\newcommand{\var}[2]{%
\ifthenelse{\equal{#2}{}}{\ProbOpr{VAR}_{#1}}
{\ifthenelse{\equal{#1}{}}{\ProbOpr{VAR}\left[#2\right]}{\ProbOpr{VAR}_{#1}\left[#2\right]}}} 

\newcommand{\vtheta}{\vct{\theta}}

\newcommand{\vp}{\vct{p}}

\newcommand{\vx}{{\vct{x}}}
\newcommand{\vy}{\vct{y}}

\newcommand{\vzero}{\vct{0}}

\newcommand{\vphi}{\vct{\phi}}

\newcommand{\vdelta}{\vct{\delta}}
\newcommand{\vomega}{\vct{\omega}}

\newcommand{\eat}[1]{}

\newcommand\id{\ensuremath{\mathbbm{1}}}
\newcommand*{\QED}{\hfill\ensuremath{\square}}

\title{Kernel Approximation Methods for Speech Recognition}

\author{Avner May$^{1\dagger}$, Alireza Bagheri Garakani$^{2\ddagger}$, Zhiyun Lu$^{2\ddagger}$, Dong Guo$^{2\ddagger}$, Kuan Liu$^{2\ddagger}$, \\
Aur\'{e}lien Bellet$^3$, Linxi Fan$^4$, Michael Collins$^{1}$\thanks{On leave at Google Inc. New York.}, Daniel Hsu$^1$, Brian Kingsbury$^5$, \\
Michael Picheny$^5$, Fei Sha$^2$ \vspace{0.1in}\\
$^1$Dept. of Computer Science, Columbia University, New York, NY 10027, USA \\
\texttt{\{avnermay, mcollins, djhsu\}@cs.columbia.edu}, \texttt{lf2422@columbia.edu} \vspace{0.1in} \\
$^2$Dept. of Computer Science, University of Southern California, Los Angeles, CA 90089, USA \\
\texttt{\{bagherig, zhiyunlu, dongguo, kuanl, feisha\}@usc.edu}\vspace{0.1in}\\
$^3$INRIA, 40 Avenue Halley, 59650 Villeneuve d'Ascq, France \\
\texttt{aurelien.bellet@inria.fr}\vspace{0.1in}\\
$^4$Dept. of Computer Science, Stanford University, Stanford, CA 94305, USA \\
\texttt{jimfan@cs.stanford.edu}\vspace{0.1in}\\
$^5$IBM T. J. Watson Research Center, Yorktown Heights, NY 10598, USA \\
\texttt{\{bedk, picheny\}@us.ibm.com}\vspace{0.1in}\\
{\emph{$^\dagger$\ $^\ddagger$: Contributed equally as the first and second co-authors, respectively}}
}
\date{}

\begin{document}

\maketitle

\begin{abstract}
We study large-scale kernel methods for acoustic modeling in speech recognition and compare 
their performance to deep neural networks (DNNs).  We perform experiments on four 
speech recognition datasets, including the TIMIT and Broadcast News benchmark 
tasks, and compare these two types of models on frame-level performance 
metrics (accuracy, cross-entropy), as well as on recognition metrics 
(word/character error rate). In order to scale kernel methods to these large 
datasets, we use the random Fourier feature method of \citet{rahimi07random}.
We propose two novel techniques for improving the performance of kernel 
acoustic models. First, in order to reduce the number of random features 
required by kernel models, we propose a simple but effective method for 
feature selection. The method is able to explore a large number of non-linear 
features while maintaining a compact model more efficiently than existing 
approaches. Second, we present a number of frame-level metrics which 
correlate very strongly with recognition performance when computed on the 
heldout set; we take advantage of these correlations by monitoring these 
metrics during training in order to decide when to stop learning.  This
technique can noticeably improve the recognition performance of both DNN and kernel 
models, while narrowing the gap between them. Additionally, we show that the 
linear bottleneck method of \citet{sainath2013low} improves the performance 
of our kernel models significantly, in addition to speeding up training and 
making the models more compact.  Together, these three methods dramatically 
improve the performance of kernel acoustic models, making their performance 
comparable to DNNs on the tasks we explored.

\end{abstract}

\begin{keywords}
Kernel Methods, Deep Neural Networks, Acoustic Modeling, Automatic Speech Recognition, Feature Selection, Logistic Regression.
\end{keywords}

\section{Introduction}
\label{sec:intro}
In recent years, deep learning techniques have significantly advanced 
state-of-the-art performance in automatic speech recognition (ASR), achieving large 
drops in word error rates \citep{seide11feature,hinton12deep, 
mohamed12dbn,parity2016}. Deep neural networks (DNNs) are able to gracefully scale to 
very large datasets, and can successfully leverage this additional data to 
achieve strong empirical performance.  In stark contrast, kernel methods, 
which are attractive due to their powerful modeling of highly nonlinear data, 
as well as for their theoretical learning guarantees and tractability 
\citep{scholkopf02}, do not scale well.  In particular, with data sets of size $N$, 
the $\Theta(N^2)$ size of the kernel matrix makes training prohibitively 
slow, while the typical $\Theta(N)$ size of the resulting models \citep{ 
steinwart2004sparseness} makes their deployment impractical.

Much recent effort has been devoted to the development of approximations to 
kernel methods, primarily via the Nystr\"om approximation \citep{nystrom} or 
via random feature expansion \citep[e.g.,][]{rahimi07random,kar12}. These 
methods yield explicit feature representations on which linear learning 
methods can provide good approximations to the original non-linear kernel 
method. However, there have been very few successful applications of these 
methods to ASR, let alone any ``head-on'' comparisons to DNNs, except for a 
few efforts which were limited in scope \citep{deng12convex,cheng11arcos, 
huang14kernel}.

In this paper, we investigate empirically how kernel methods can be scaled to 
tackle typical ASR tasks. We focus on four datasets: the IARPA Babel Program 
Cantonese (IARPA-babel101-v0.4c) and Bengali (IARPA-babel103b-v0.4b) limited 
language packs, a 50-hour subset of Broadcast News (BN-50) 
\citep{kingsbury09,sainath2011making}, and TIMIT \citep{timit}. We
present several results: First, we show that kernel methods can 
be efficiently scaled to large-scale ASR tasks, using the above-mentioned random 
Fourier feature technique \citep{rahimi07random}. Our contribution is to 
demonstrate the practical utility of this method in constructing large-scale 
classifiers for acoustic modeling.  Second, we have found that when 
leveraging the novel techniques discussed in this paper, our kernel-based 
acoustic models are generally competitive with layer-wise discriminatively 
pre-trained DNN-based models \citep{seide11pretrain}.

In order to attain strong performance for the kernel acoustic models, we have developed 
a few new methods. First, we propose a simple feature selection algorithm, 
which effectively reduces the number of random features required. 
We iteratively select features from large pools of 
random features, using learned weights in the selection criterion.  This has 
two clear benefits: (i) the subsequent training on the selected features is 
considerably faster than training on the entire pool of random features, and
(ii) the resulting model is also much smaller. For certain kernels, this
feature selection approach---which is applied at the level of the random 
features---can be regarded as a non-linear method for feature selection at 
the level of the input features, and we use this observation to motivate 
the design of a new kernel function.

Second, we present several novel frame-level metrics which correlate very 
strongly with the token error rate (TER),\footnote{For our Cantonese dataset, 
`token error rate' corresponds to `character error rate.'  For our Bengali and 
Broadcast News datasets, it corresponds to `word error rate.'  For TIMIT, it 
corresponds to `phone error rate.'} and which can thus be monitored on the 
heldout set during training in order to determine when to stop learning.  
Using this method, we achieve notable gains in TER for both kernels and 
DNNs.  This method partially mitigates a well-known problem in acoustic 
modeling; namely, that the training criterion (cross-entropy) often does not 
align well with the true objective (TER). In our case, we noticed that although
our kernel and DNN models 
would often attain very similar cross-entropy values on the heldout set, the 
DNNs would generally perform better, sometimes by a wide margin, in terms of 
TER. Although sequence training techniques can also be used to address this 
issue \citep[e.g.,][]{kingsbury09,vesely13}, they are very computationally 
expensive, and they generally depend on frame-level training for initialization; thus, 
our proposed method can be used in conjunction with existing sequence 
training techniques, by providing them with a better initial model.

Lastly, we demonstrate the importance of using a linear bottleneck 
\citep{sainath2013low} in the parameter matrix of our kernel models.  Not only does 
this method improve the performance of our kernel models significantly, it 
also makes training faster, and reduces the size of the models learned.

This paper builds on the previous works of \citet{lu2016} and \citet{may2016}.
\citet{lu2016} provide comparisons between DNN and kernel acoustic models 
on 3 datasets (Cantonese, Bengali, and Broadcast News); they additionally 
present the ``entropy regularized log loss'' (ERLL) metric, and show how 
using it as a model selection criterion can yield TER improvements.\footnote{In
\citet{lu2016} this metric was called ``entropy regularized perplexity'' (ERP).}
\citet{may2016} present the feature selection algorithm described in this paper, 
along with ASR experiments on two datasets (Cantonese and Bengali); 
comparisons to DNNs are also performed. The work in the current paper builds 
on this existing work in several ways.  First, we provide a more extensive 
set of experiments, including results on the ASR benchmark TIMIT dataset, in 
addition to updated results on the other datasets (Cantonese, Bengali, and 
Broadcast News). Second, we have extended the work on ERLL by presenting a 
larger set of metrics which correlate strongly with TER; these additional 
metrics help explain the unusual correlation between ERLL and TER, as well as 
the poor correlation between cross-entropy and TER.  In this paper, we show 
how these metrics can be evaluated on the heldout set during training in 
order to decide when to decay the learning rate and stop training.  Lastly, 
we provide an extensive set of experiments showing the importance of using a 
linear bottleneck for attaining strong TER performance for our kernel methods.

The rest of the paper is organized as follows. We review related work in 
\S\ref{sec:relwork}. We provide some background for kernel approximation 
methods, as well as for acoustic modeling, in \S\ref{sec:background}.  We 
present our feature selection algorithm in \S\ref{sec:method_featsel}. In 
\S\ref{sec:method_earlystop}, we present several novel metrics which 
correlate strongly with TER, and show how they can be used during training to 
improve TER performance. In \S\ref{sec:experiments}, we report extensive 
experiments comparing DNNs and kernel methods, including results using the 
methods discussed above. We conclude in \S\ref{sec:conclusion}.

\section{Related Work}
\label{sec:relwork}
Scaling up kernel methods has been a long-standing and actively studied 
problem \citep{largescalekernelmachines07, smolaXX, decoste02invariant, 
platt98smo, tsang05cvm, clarkson10coreset}. For kernels with sparse feature 
expansions, \citet{sonnenburg10coffin} show how to efficiently scale kernel 
SVMs to datasets with up to 50 million training samples by using sparse
vector operations for parameter updates.  Approximating 
kernels by constructing explicit finite-dimensional feature representations, 
where dot products between these representations approximate the kernel 
function, has emerged as a powerful technique \citep[e.g.,][]{nystrom, 
rahimi07random}. The Nystr\"om method constructs these feature maps, 
for arbitrary kernels, via a low-rank decomposition of the kernel matrix 
\citep{nystrom}. For shift-invariant kernels, the random Fourier feature 
technique of \citet{ rahimi07random} uses random projections in order to 
generate the features. Random projections can also be used to approximate a 
wider range of kernels \citep{kar12, vedaldi12additive, hamid14, pennington15}.
Many recent works have been developed to speed-up the random Fourier feature 
approach to kernel approximation.  One line of work attempts to reduce the 
time (and memory) needed to compute the random feature expansions by imposing 
structure on the random projection matrix \citep{fastfood,yu15}. It is also 
possible to use doubly-stochastic methods to speed-up stochastic gradient 
training of models based on the random features \citep{dai14}. 

Despite much progress in kernel approximation, there have 
been only a few reported empirical studies of these techniques on speech 
recognition tasks \citep{deng12convex,cheng11arcos,huang14kernel}. However, 
those tasks were relatively small-scale (for instance, on the TIMIT dataset). 
For the most part, a detailed evaluation of these methods on large-scale ASR 
tasks, together with a thorough comparison with DNNs, is lacking. Our work 
fills this gap, tackling challenging large-scale acoustic modeling problems, 
where deep neural networks achieve strong performance \citep{hinton12deep,
dahl2012context}.  Additionally, we provide a number of important improvements 
to the kernel methods, which boost their performance significantly.

One contribution of our work is to introduce a feature selection method that 
works well in conjunction with random Fourier features in the context of 
large-scale multi-class classification problems. Recent work on feature 
selection methods with random Fourier features, for binary classification and 
regression problems, includes the Sparse Random Features algorithm of 
\citet{sparseRKS}. This algorithm is a coordinate 
descent method for smooth convex optimization problems in the (infinite) 
space of non-linear features: each step involves solving a batch
$\ell_1$-regularized convex optimization problem over randomly generated non-linear
features (note that a natural extension of this method to multi-class 
problems is to use mixed norms such as $\ell_1/\ell_2$).  Here, the
$\ell_1$-regularization may cause the learned solution to only depend on a subset of
the generated features. A drawback of this approach is the computational 
burden of fully solving many batch optimization problems, which is 
prohibitive for large data sets.  In our attempts to implement an online 
variant of this method, using FOBOS \citep{fobos} and 
$\ell_1/\ell_2$-regularization for the multi-class setting, we observed that very strong 
regularization was required to obtain any intermediate sparsity, which in 
turn severely hurt prediction performance. Effectively, the regularization 
was so strong that it made the learning meaningless, and the selected 
features were basically random.  Our approach for selecting random features 
is more efficient, and more directly ensures sparsity,
than regularization.

Another improvement we propose alters the frame-level training of the 
acoustic model in order to improve the recognition performance (TER) of the 
final model.  A set of methods, typically referred to as \textit{sequence 
training} techniques, share our goal of tuning the acoustic model for the 
purpose of improving its recognition performance.  There are a number of 
different sequence training criteria which have been proposed, including 
maximum mutual information (MMI) \citep{bahl86,valtchev97}, boosted MMI (BMMI)
 \citep{povey08}, minimum phone error (MPE) \citep{povey02}, or minimum Bayes 
risk (MBR) \citep{kaiser00,gibson06,povey07}.  These methods, though 
originally proposed for training Gaussian mixture model (GMM) acoustic 
models, can also be used for neural network acoustic models \citep{kingsbury09,
vesely13}. Nonetheless, all of these methods are quite computationally 
expensive and are typically 
initialized with an acoustic model trained via the frame-level cross-entropy 
criterion. Our method, by contrast, is very simple, only making a small 
change to the frame-level training process.  Furthermore, it can be used in 
conjunction with the above-mentioned sequence training techniques, by 
providing a better initial model.  Recently, \citet{povey16} showed that it 
is possible to train an acoustic model using \textit{only} sequence-training 
methods, with the lattice-free version of the MMI criterion. For future work, 
we would like to see how much our kernel models can benefit from the various 
sequence training methods mentioned above, relative to DNNs.

This work also contributes to the debate on the relative strengths of deep 
and shallow neural networks.  As explained in Section 
\ref{subsec:kernel-shallow}, many types of kernels (including popular kernels
like the Gaussian kernel and the Laplacian kernel) can be understood as
shallow neural networks. As such, comparing kernel methods to 
DNNs is also in a sense comparing shallow and deep neural networks. There is 
much literature on this topic.  Classic results show that both deep and 
shallow neural networks are ``universal approximators,'' meaning that they 
can approximate any real-valued continuous function with bounded support to 
an arbitrary degree of precision \citep{cybenko89,hornik89}. However, a 
number of papers have argued that there exist functions which deep neural 
networks can express with exponentially fewer parameters than shallow neural 
networks \citep{montufar14,bianchini14}.  In \citet{ba14}, the authors show 
that the performance of shallow neural networks can be increased considerably 
by training them to match the outputs of deep neural networks.
In showing that kernel methods can compete with DNNs on large-scale speech 
recognition tasks, this paper adds credence to the argument that shallow 
networks can perform on par with deep networks.

\section{Background}
\label{sec:background}
\subsection{Kernel Methods and Random Features}
\label{subsec:RFF}

Kernel methods, broadly speaking, are a set of machine learning techniques 
which either explicitly or implicitly map data from the input space $\cX$ to 
some feature space $\cH$, in which a linear model is learned. A ``kernel 
function'' $\kernel\colon\cX \times \cX \rightarrow \R$ is then 
defined\footnote{It is also possible to define the kernel function prior to 
defining the feature map; then, for positive-definite kernel functions, Mercer's 
theorem guarantees that a corresponding feature map $\vphi$ exists such that
$\kernel(\vx,\vy) =   \Dotp{\vphi(\vx),\vphi(\vy)}$.} as the function which 
takes as input $\vx,\vy \in \cX$, and returns the dot-product of the 
corresponding points in $\cH$.  If we let $\vphi\colon\cX \rightarrow \cH$ 
denote the map into the feature space, then 
$\kernel(\vx,\vy) = \Dotp{\vphi(\vx),\vphi(\vy)}$.  
Standard kernel methods avoid inference in $\cH$, because 
it is generally a very high-dimensional, or even infinite-dimensional, space. 
Instead, they solve the dual problem by using the $N$-by-$N$ kernel matrix, 
containing the values of the kernel function applied to all pairs of $N$ training 
points. When $\dim(\cH)$ is far greater than $N$, this ``kernel trick'' provides
a nice computational advantage. However, when $N$
is exceedingly large, the $\Theta(N^2)$ size of the kernel matrix makes 
training impractical.

\citet{rahimi07random} address this problem by leveraging Bochner's Theorem, 
a classical result in harmonic analysis, in order to provide a fast way to 
approximate any positive-definite \emph{shift-invariant} kernel $\kernel$ 
with \emph{finite}-dimensional features.  A kernel $\kernel(\vx,\vy)$ is 
\emph{shift-invariant} if and only if $\kernel(\vx,\vy) = \hat{\kernel}(\vx-\vy)$ 
for some function $\hat{\kernel}\colon\R^d\rightarrow \R$.  We now present 
Bochner's Theorem:

\begin{theorem}{(Bochner's theorem, adapted from \cite{rahimi07random})}: 
A continuous shift-invariant kernel $\kernel(\vx,\vy) = \hat{\kernel}(\vx-\vy)$
on $\R^d$ is positive-definite if and only if $\hat{\kernel}$ is the Fourier
transform of a non-negative measure $\mu(\vomega)$.
\end{theorem}
Thus, for any positive-definite shift-invariant kernel $\hat{\kernel}(\vdelta)$, we have that
\begin{equation}
\hat{\kernel}(\vdelta) = \int_{\R^d} \mu(\vomega) e^{-j\vomega\T\vdelta}\, d\vomega,\\
\label{eq:FT}
\end{equation}
where
\begin{equation}
\mu(\vomega) = (2\pi)^{-d}\int_{\R^d} \hat{\kernel}(\vdelta) e^{j\vomega\T\vdelta}\, d\vdelta
\label{eq:IFT}
\end{equation}
is the inverse Fourier transform\footnote{There are various ways of
  defining the Fourier transform and its inverse.  We use the
  convention specified in Equations \eqref{eq:FT} and \eqref{eq:IFT},
  which is consistent with \citet{rahimi07random}.} of
$\hat{\kernel}(\vdelta)$, and where $j=\sqrt{-1}$. By Bochner's
theorem, $\mu(\vomega)$ is a non-negative measure. As a result, if we
let $Z = \int_{\R^d} \mu(\vomega) d\vomega$, then
$p(\vomega)=\frac{1}{Z}\mu(\vomega)$ is a proper probability
distribution, and we get that
\begin{equation*}
\frac{1}{Z}\hat{\kernel}(\vdelta) = \int_{\R^d} p(\vomega) e^{-j\vomega\T\vdelta}\, d\vomega.\\
\end{equation*}

For simplicity, we will assume going forward that $\hat{\kernel}$ is 
properly-scaled, meaning that $Z=1$.  Now, the above equation allows us to 
rewrite this integral as an expectation:

\begin{equation}
\hat{\kernel}(\vdelta) = \hat{\kernel}(\vx - \vy)  = \int_{\R^d} p(\vomega) e^{j\vomega\T(\vx-\vy)}\, d\vomega = \expect{\vomega}{e^{j\vomega\T\vx}e^{-j\vomega\T\vy}}.
\label{eq:RFF1}
\end{equation}
This can be further simplified as
\begin{equation*}
\hat{\kernel}(\vx - \vy)  = \expect{\vomega,b}{\sqrt{2}\cos(\vomega\T\vx + b)\cdot \sqrt{2}\cos(\vomega\T\vy + b)},
\label{eq:RFF2}
\end{equation*}
where $\vomega$ is drawn from $p(\vomega)$, and $b$ is drawn uniformly
from $[0,2\pi]$.  See Appendix~\ref{sec:appendixA} for details on why
this specific functional form is correct.

This motivates a sampling-based approach for approximating the kernel 
function. Concretely, we draw $\{\vomega_1, \vomega_2, \ldots, \vomega_D\}$ 
independently from the distribution $p(\vomega)$, and $\{b_1,b_2,\ldots\,b_D\}$ 
independently from the uniform distribution on $[0,2\pi]$, and then use these
parameters to approximate the kernel, as follows:

\begin{equation*}
\kernel(\vx, \vy) \approx \frac{1}{D} \sum_{i=1}^{D} \sqrt{2}\cos(\vomega_i\T\vx + b_i)\cdot \sqrt{2}\cos(\vomega_i\T\vy + b_i) = z(\vx)\T z(\vy),
\end{equation*}
where $z_i(\vx) = \sqrt{\frac{2}{D}}\cos(\vomega_i\T\vx + b_i)$ is the $i^{th}$
element of the $D$-dimensional random vector $z(\vx)$.
In Table \ref{table:GaussLap}, we list two popular (properly-scaled) 
positive-definite kernels with their respective inverse Fourier transforms.

\begin{table}
\centering
\begin{tabular}{|c|c|c|c|}\hline
\textbf{Kernel name} & $\bm{\kernel(x,y)}$ & $\bm{p(\omega)}$ & \textbf{Density name} \\ \hline
Gaussian & $e^{-\|\vx-\vy\|_2^2/2\sigma^2}$ 
& $(2\pi (1/\sigma^2))^{-d/2}e^{-\frac{\|\vomega\|_2^2}{2(1/\sigma)^2}}$  & $\text{Normal}(\vzero_d,\frac{1}{\sigma^2}\id_d)$ \\ \hline
Laplacian & $e^{-\lambda\|\vx-\vy\|_1}$ & $\prod_{i=1}^d\frac{1}{\lambda\pi(1 + (\vomega_i/\lambda)^2)}$  
& $\text{Cauchy}(\vzero_d,\lambda)$ \\ \hline
\end{tabular}
\caption{Gaussian and Laplacian Kernels, together with their sampling distributions $p(\vomega)$}
\label{table:GaussLap}
\end{table}

Using these random feature maps, in conjunction with linear learning 
algorithms, can yield huge gains in efficiency relative to standard kernel 
methods on large datasets. Learning with a representation $z(\cdot) \in \R^D$ 
is relatively efficient provided that $D$ is far less than the number of 
training samples $N$.  For example, in our experiments (see Section
\ref{sec:experiments}), we have $2$ million to $16$ million training samples, 
while $D \approx \num[group-separator={,}]{25000}$ often leads to good performance.

\citet{rahimi07random,rahimi08kitchen} prove a number of important theoretical results about these
random feature approximations.  First, they show that if 
$D=\tilde\Omega(\frac{d}{\epsilon^2})$, then with high probability $z(\vx)\T z(\vy)$ will be 
within $\epsilon$ of $\kernel(\vx,\vy)$ for all $\vx,\vy$ in some compact 
subset $\cM\in \R^d$ of bounded diameter.\footnote{We are using the 
$\tilde{\Omega}$ notation to hide logarithmic factors.}  See claim 1 of 
\citet{rahimi07random} for the more precise statement and proof of this result.

In their follow-up work \citep{rahimi08kitchen}, the authors prove a 
generalization bound for models learned using these random features. They 
show that with high-probability, the excess risk\footnote{The ``risk'' of a 
model is defined as its expected loss on unseen data.} assumed from using this 
approximation, relative to using the ``oracle'' kernel model (the exact 
kernel model with the lowest risk), is bounded by 
$O(\frac{1}{\sqrt{N}} + \frac{1}{\sqrt{D}})$ (see the main result of 
\citet{rahimi08kitchen} for more details).
Given that the generalization error of a model trained using exact kernel 
methods is known to be within $O(\frac{1}{\sqrt{N}})$ of the oracle model 
\citep{bartlett2002}, this implies that in the worst case, $D=\Theta(N)$ 
random features may be required in order for the approximated model to 
achieve generalization performance comparable to the exact kernel model.  
Empirically, however, fewer than $\Theta(N)$ features are often needed in 
order to attain strong performance \citep{yu15}.

\subsection{Using Neural Networks for Acoustic Modeling}
\label{subsec:DNN-acoustic-modeling}

Neural network acoustic models provide a conditional probability distribution 
$p(y|\vx)$ over $C$ possible acoustic states, conditioned on an acoustic 
frame $\vx$ encoded in some feature representation. The acoustic states 
correspond to context-dependent phoneme states \citep{dahl2012context}, and 
in modern speech recognition systems, the number of such states is of the 
order $10^3$ to $10^4$. The acoustic model is used within probabilistic 
systems for decoding speech signals into word sequences.  Typically, the 
probability model used is a hidden Markov model (HMM), where the model's 
emission and transition probabilities are provided by an acoustic model 
together with a language model. We use Bayes' rule in order to compute the 
probability $p(\vx | y)$ of emitting a certain acoustic feature vector $\vx$ 
from state $y$, given the output $p(y|\vx)$ of the neural network:
\begin{eqnarray*}
p(\vx | y) &=& \frac{p(y|\vx)p(\vx)}{p(y)}.
\end{eqnarray*}
Note that $p(x)$ can be ignored at inference time because it doesn't affect 
the relative scores assigned to different word sequences, and $p(y)$ is 
simply the prior probability of HMM state $y$. The Viterbi algorithm can then 
be used to determine the most likely word sequence (see \citet{gales2007} for 
an overview of using HMMs for speech recognition).

\subsection{Using Random Fourier Features for Acoustic Modeling}
\label{subsec:RFF-acoustic-modeling}

In order to train an acoustic model using random Fourier features, we can 
simply plug the random feature vector $z(\vx)$ (for an acoustic frame $\vx$) 
into a multinomial logistic regression model:

\begin{equation}
p( y |\vx) = \frac{\exp\big(\Dotp{\vtheta_y, \;z(\vx)}\big)}{\sum_{y'}\exp\big(\Dotp{\vtheta_{y'},\; z(\vx)}\big)}.
\label{eq:MLR}
\end{equation}
The label $y$ can take any value in $\{1, 2, \ldots, C\}$, each corresponding 
to a context-dependent phonetic state label, and the parameter matrix 
$\Theta = [\vtheta_1|\ldots|\vtheta_C]$ is learned. Note that we also include a bias 
term, by appending a 1 to $z(\vx)$ in the equation above.

\subsection{Viewing Kernel Acoustic Models as Shallow Neural Networks}
\label{subsec:kernel-shallow}

\begin{figure}
\centering
\includegraphics[width=0.5\columnwidth]{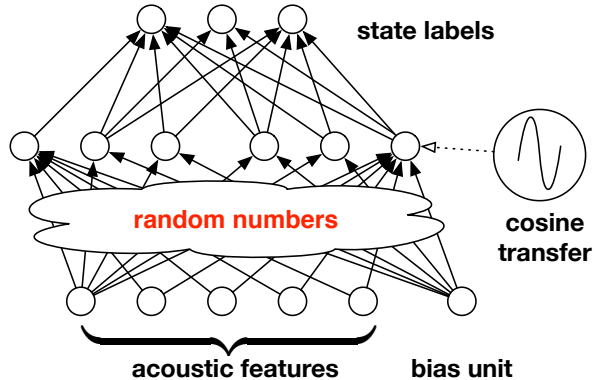}
\vspace{-1em}
\caption{Kernel-acoustic model seen as a shallow neural network}
\label{fig:shallow}
\vspace{-1em}
\end{figure}

The model in Equation \eqref{eq:MLR} can be seen as a shallow neural network, 
with the following properties: (1) the parameters from the inputs (\ie, 
acoustic feature vectors) to the hidden units are set randomly, and are not 
learned; (2) the hidden units use $\cos(\cdot)$ as their activation function; 
(3) the parameters from the hidden units to the output units are learned (can 
be optimized with convex optimization); and (4) the softmax function is used 
to normalize the outputs of the network.  See Figure~\ref{fig:shallow} for a 
visual representation of this model architecture.

\subsection{Linear Bottlenecks}
\label{background:bottleneck}
The number of phonetic state labels can be very large. This will 
significantly increase the number of parameters in $\Theta$. We can reduce 
this number with a \textit{linear bottleneck} layer between the hidden layer 
and the output layer; the linear bottleneck corresponds to a low-rank 
factorization $\Theta = UV$ of the parameter matrix \citep{sainath2013low}. 
This is particularly important for our kernel 
models, where the number of trainable parameters is $D\times C$, where 
$D$ is the number of random features, and $C$ is the number of output classes.  
Using a linear bottleneck of size $r$, this can be reduced to 
$(D\times r + r\times C)$, which is significantly less than 
$D\times C$ when $r \ll \min(D,C)$.  
This strictly decreases the capacity of the resulting model, while unfortunately
rendering the optimization problem non-convex.

\section{Random Feature Selection}
\label{sec:method_featsel}
In this section, we first motivate and describe our proposed feature 
selection algorithm.  We then introduce a new ``Sparse Gaussian'' kernel, 
which performs well in conjunction with the feature selection algorithm.

\subsection{Proposed Feature Selection Algorithm}
\label{sec:method-feature-selection}

Our proposed random feature selection method, shown in Algorithm~\ref{alg:rfs},
is based on a general iterative framework. In each iteration, random 
features are generated and added to the current set of features; a subset of
these features are selected, while the rest are discarded. The selection process works as 
follows: first, a model is trained on the current set of features using a 
single pass of stochastic gradient descent (SGD) on a subset of the training 
data.  Then, the features whose corresponding rows in $\Theta$ have the 
largest $\ell_2$ norms are kept.  Note that the row of weights corresponding to
the feature $z_i(\vx)$ in the model $f(\vx) = h(\Theta\T z(\vx))$ are those in 
the $i^{th}$ row of $\Theta$. In the case where we are using a linear 
bottleneck to decompose $\Theta$ into $UV$, we perform the SGD 
training using this decomposition. After we complete the training in a given 
iteration, we compute $\Theta=UV$, and then select features based on the 
$\ell_2$-norms of the rows of $\Theta$.

This feature selection method has the following advantages: The overall 
computational cost is mild, as it requires just $T$ passes through subsets of 
the data of size $R$ (equivalent to $\sim TR/N$ full SGD epochs). In fact, in 
our experiments, we find it sufficient to use $R=O(D)$. 
Moreover, the method is able to explore a large number of 
non-linear features, while maintaining a compact model. If $s_t = Dt/T$, then 
the learning algorithm is exposed to $D(T+1)/2$ random features 
throughout the feature selection process; this is the selection schedule we used
in all our experiments. We show in Section 
\ref{sec:experiments} that this empirically increases the predictive quality of 
the selected non-linear features.

It is important to note the similarities between this method, and the FOBOS 
method with $\ell_1/\ell_2$-regularization \citep{fobos}.  In the latter 
method, one solves the $\ell_1/\ell_2$-regularized problem in a stochastic 
fashion by alternating between taking unregularized stochastic gradient 
descent (SGD) steps, and then ``shrinking'' the rows of the parameter matrix; 
each time the parameters are shrunk, the rows whose $\ell_2$-norms are below a 
threshold are set to $0$.  After training completes, the solution will likely 
have some rows which are all zero, at which point the features corresponding 
to those rows can be discarded. In our method, on the other hand, we take 
many consecutive unregularized SGD steps, and only thereafter do we choose to 
discard the rows whose $\ell_2$-norm is below a threshold.  As mentioned in 
the Related Work section, our attempts at using FOBOS for feature selection 
failed, because the magnitude of the regularization parameter needed in order to produce a 
sparse model was so large that it dominated the learning process; as a 
result, the models learned performed badly, and the selected features were 
essentially random.

One disadvantage of our method is that the index used for selection may 
misrepresent the features' actual predictive utilities. For instance, the 
presence of some random feature may increase or decrease the weights for 
other random features relative to what they would be if that feature were not
present. An alternative would be to consider features 
in isolation, and add features one at a time (as in stagewise regression 
methods and boosting), but this would be significantly more computationally 
expensive.  For example, it would require $O(D)$ passes through the data, 
relative to $O(T)$ passes, which would be prohibitive for large $D$ values. 
We find empirically that the influence of the additional random features in the 
selection criterion is tolerable, and it is still possible to select useful 
features with this method.

\begin{algorithm}[t]
  \caption{%
    Random feature selection%
  }
  \label{alg:rfs}
  \begin{algorithmic}[1]
    \renewcommand{\algorithmicrequire}{\textbf{input}}
    \REQUIRE
    Target number of random features $D$, data subset size $R$, \\
    Integers $(s_1,\ldots,s_{T-1})$ such that $0 < s_1 < \dotsb < s_{T-1} < D$,
		specifying selection schedule.
		\STATE \textbf{initialize} set of selected indices  $S := \emptyset$.
    \FOR{$t=1,2,\dotsc,T$}
			\FOR{$i \in \{1,\ldots,D\} \backslash S$}
				\STATE $\vomega_i \sim p(\vomega)$.
				\STATE $b_i \sim \cU(0,2\pi)$.
			\ENDFOR
			\IF{$t \neq T$}
			\STATE Initialize parameter matrix $\Theta$.\footnotemark
			\STATE Learn weights $\Theta \in \RR^{D \times C}$ using a single pass 
of SGD over $R$ randomly selected training examples, using the
projection vectors $(\vomega_1,\ldots,\vomega_D)$, and the
biases $(b_1,\ldots,b_D)$, to generate the random Fourier features.
			\STATE $S := \{i \;|\; \Theta_i \text{ is amongst the } s_t 
					\text{ rows of } \Theta \text{ with highest }\ell_2 \text{ norm}\}.$
			\ENDIF
		\ENDFOR
    \RETURN The selected projection vectors $(\vomega_1,\ldots,\vomega_D)$, and 
		        the selected biases $(b_1,\ldots,b_D)$.
  \end{algorithmic}
\end{algorithm}
\footnotetext{See Section \ref{sec:exp_train_details} for details on how
$\Theta$ is initialized.}
\subsection{A Sparse Gaussian Kernel}
\label{sec:method-sparse-rbf}

Recall from Table~\ref{table:GaussLap} that for the Laplacian kernel, the 
sampling distribution used for the random Fourier features is the 
multivariate Cauchy density $p(\vomega) \propto \prod_{i=1}^d (1+\omega_i^2)^{-1}$
(we let $\lambda=1$ here for simplicity). If we draw
$\vomega = (\omega_1,\ldots,\omega_d)$ from $p$, then each $\omega_i$ has a 
two-sided fat tail distribution, and hence $\vomega$ will typically contain some 
entries much larger than the rest.

This property of the sampling distribution implies that many of the random 
features generated in this way will effectively concentrate on only a few of 
the input features. We can thus regard such random features as being 
non-linear combinations of a small number of the original input features. 
Thus, the proposed feature selection method 
effectively picks out useful non-linear interactions between small sets of 
input features.

We can also directly construct sparse non-linear combinations of the input 
features.  Instead of relying on the properties of the Cauchy distribution, 
we can actually choose a small number $k$ of coordinates 
$F \subseteq \{1,2,\dotsc,d\}$, say, uniformly at random, and then choose the random vector 
$\vomega$ so that it is always zero in positions outside of $F$; the same 
non-linearity (e.g., $\vx \mapsto \cos(\vomega \T \vx +b)$) can be applied once 
the sparse random vector is chosen.  Compared to the random Fourier feature 
approximation to the Laplacian kernel, the vectors $\vomega$ chosen in this 
way are truly sparse, which can make the random feature expansion more 
computationally efficient to apply (if efficient sparse matrix operations are 
used).

Note that random Fourier features with such sparse sampling distributions in 
fact correspond to shift-invariant kernels that are rather different from the 
Laplacian kernel. For instance, if the non-zero entries of $\vomega$ are 
drawn i.i.d. from $\cN(0,\sigma^{-2})$, then the corresponding kernel is

\begin{equation}
  \label{eq:sparse-gauss-kernel}
  \kernel(\vx,\vy)
  =
  \binom{d}{k}^{-1}
	\sum_{
    \substack{
      F \subseteq \{1,\ldots,d\} \\ |F|=k
    }
  }
  \;\;
	\exp\Parens{-\frac{\|\vx_F-\vy_F\|_2^2}{2\sigma^2}},
\end{equation}
where $\vx_F$ is a vector composed of the elements $\vx_i$ for $i \in F$.
The kernel in Equation \eqref{eq:sparse-gauss-kernel} puts equal emphasis on all 
input feature subsets $F$ of size $k$. However, the feature selection process 
may effectively bias the distribution of the feature subsets to concentrate on 
some small family $\cF$ of input feature subsets.

\section{New Early Stopping Criteria}
\label{sec:method_earlystop}
As discussed in the introduction, there is a well-known problem in the 
training of acoustic models; namely, that the training criterion 
(cross-entropy) does not perfectly correlate with the true objective (TER).  Consequently,
lowering the cross-entropy performance on a heldout set does not necessarily 
result in better TER performance. For example, we noticed that our DNNs were 
often attaining stronger TER performance than our kernel models, even though 
they had comparable cross-entropy performance. In order to partially address 
this problem, in this section we present several new metrics whose empirical
correlation with TER, amongst the fully trained DNN and kernel models we 
trained, was high. We then leverage these metrics during training by 
evaluating them on the heldout set after each epoch in order to decide when 
to decay the learning rate and stop training (see Section \ref{sec:exp_train_details}
for details on how we decay the learning rate). Note that the reason we
use these metrics as proxies for the TER, instead of directly using the TER, is 
that it is very expensive to compute the TER on the development set.

The common thread which unites all the metrics we will present is that they 
do not penalize very incorrect examples (meaning, examples for which the 
model assigned a probability very close to 0 to the correct label) as 
strongly as cross-entropy does.  Notice, for instance, that there is no limit 
to how much the cross-entropy loss (\eg, log loss) can penalize a single 
incorrect example.  Our metrics are more lenient.  We present them now:

\begin{enumerate}
\item ``\textbf{Entropy Regularized Log Loss (ERLL)}:''
This loss rewards models for being confident (\eg, low entropy), by 
considering a weighted sum of the cross entropy loss (CE) and the average 
entropy (ENT) of the model on the heldout data. Specifically, for any $\beta 
\in \R$ (typically we take $\beta=1$), we define the loss as follows:

\begin{equation*}
CE + \beta \cdot ENT = -\frac{1}{N}\sum_{i=1}^N \sum_{y=1}^C
[\mathbb{I}(y = y_i) + \beta \cdot p(y|x_i)]\log p(y|x_i)
\end{equation*}

This metric encourages models to be more confident, even if it means having a 
worse cross-entropy loss as a result.

\item ``\textbf{Capped Log Loss}:''  For any value of $\lambda \geq 0$, we can define:
\begin{equation*}
-\frac{1}{N} \sum_{i=1}^N \log (p(y_i | \vct{x}_i) + \lambda).
\end{equation*}

Effectively, this loss ensures that no single example contributes more than $-
\frac{1}{N}\log(\lambda)$ to the loss.  If $\lambda$ is a small positive 
number, this loss is very similar to the normal log loss for values of $p(y_i 
| \vct{x}_i)$ close to 1, while affecting the loss dramatically for values 
close to 0 (for example, when $p(y_i | \vct{x}_i) < \lambda$).

\item ``\textbf{Top-k Log Loss}:'' For this loss, assume that the heldout 
examples $(\vct{x}_i,y_i)$ are sorted in descending order of their $p(y_i | 
\vct{x}_i)$ values.  Now, for any positive integer $k \leq N$, we can 
define the ``Top-k Log Loss'' as follows:

\begin{equation*}
-\frac{1}{k} \sum_{i=1}^k \log p(y_i | \vct{x}_i).
\end{equation*}

This metric judges a model based on how well it does on the $k$ heldout 
examples to which it assigns highest probabilities.
\end{enumerate}

\begin{figure}
	\centering
    \begin{tabular}{ccc}
			\includegraphics[width=0.3\textwidth]{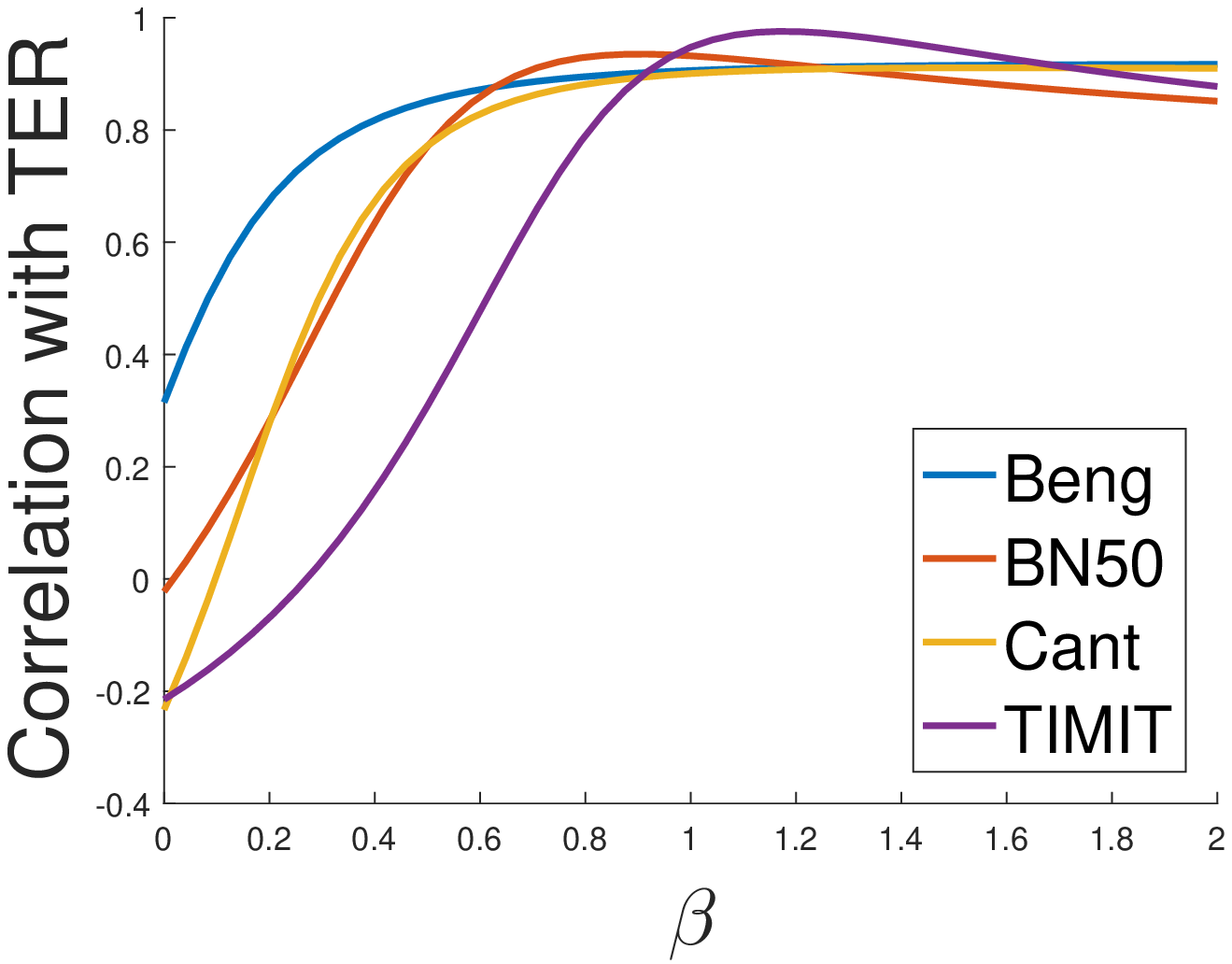} &
			\includegraphics[width=0.3\textwidth]{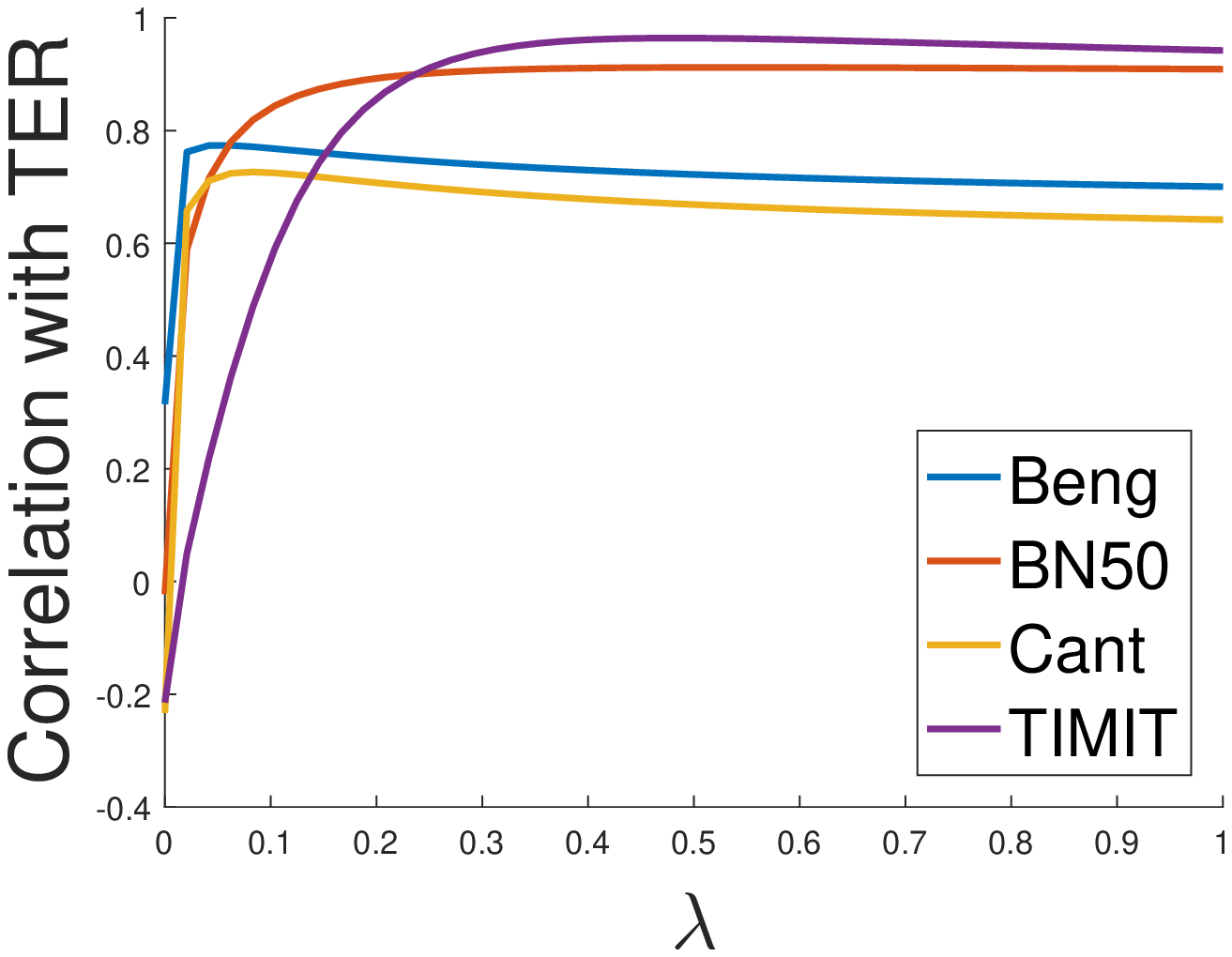} &
			\includegraphics[width=0.3\textwidth]{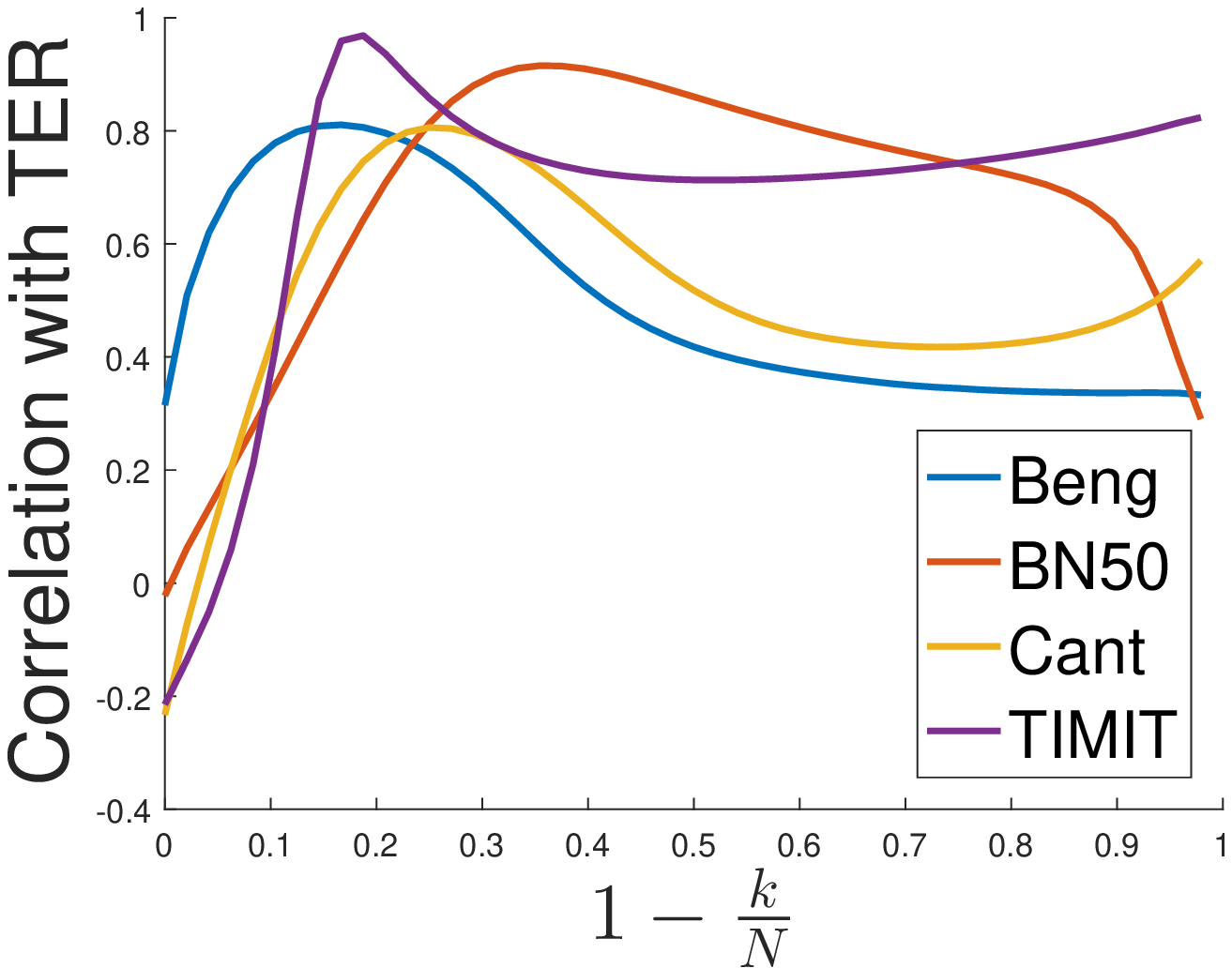}
		\end{tabular}
  \caption{%
    Empirical correlations of (left to right) Entropy Regularized Log Loss, 
		Capped Log Loss, and Top-k Log Loss with TER, as a function of $\beta$,
		$\lambda$, and $1-\frac{k}{N}$, respectively.
  }
  \label{fig:corr-with-ter}
\end{figure}

Notice that for $\beta = 0$, $\lambda = 0$, and $k=N$, these metrics all 
simplify to the standard log loss. In Figure~\ref{fig:corr-with-ter}, we show 
plots of the empirical correlations of these metrics with TER values, as a 
function of each metric's ``hyperparameters,'' based on models we have trained.  
More specifically, we \textit{fully} train a large number of kernel and DNN 
models, and then evaluate the TER performance of these models on the 
development set, as well as compute the heldout performance of these models 
in terms of the 3 metrics described above (for various settings of $\beta$,
$\lambda$, and $k$). The precise set of models we used are those in Tables 
\ref{table:TER-kernel} and \ref{table:TER-dnn} (see Section 
\ref{sec:exp_results} for details). We then compute the empirical 
correlations between these values, and plot them as a function of each 
metric's hyperparameters. Note that for the Top-k Log Loss, we plot the 
correlation with TER as a function of the fraction $1-\frac{k}{N}$ of the 
heldout dataset which is \textit{ignored}.

As can be seen from these plots, for certain ranges of values of the metric 
hyperparameters, the correlation of these metrics with TER is quite high.  
For example, for $\beta = 1$, the correlations between entropy regularized 
log loss and TER are $0.91$, $0.93$, $0.90$, and $0.95$ for Bengali, BN-50, 
Cantonese, and TIMIT respectively.  This is compared to correlations of $0.31$
, $-0.02$, $-0.23$, and $-0.21$ for the cross-entropy objective.

Based on this analysis, one reasonable thing to do would be to use the heldout 
Entropy Regularized Log Loss as a stopping criteria, instead of the standard 
cross-entropy loss. In practice, this results in the training continuing
past the point of lowest heldout cross-entropy, and producing models with lower
heldout entropy (and lower ERLL), which hopefully have a lower TER as well.
Similarly, one could  use the heldout Entropy Regularized Log Loss, instead of 
the heldout cross-entropy, in order to decide when to decay the learning rate.
This is what we do in our experiments.  The results are reported in Section 
\ref{sec:experiments}.  We use $\beta = 1$ in all our experiments.  We note that
we could have also used the other metrics (Capped, Top-k) for this 
purpose, but chose to use ERLL with $\beta=1$, since we observed that it 
attained high correlation values with TER on models we trained, across all 4 
datasets (see Figure~\ref{fig:corr-with-ter}).

\section{Experiments}
\label{sec:experiments}
In this section, we first provide a description of the datasets we use, and 
our evaluation criteria. We then give an overview of our training procedure, 
and provide details regarding hyperparameter choices.  We then present our 
experimental results comparing the performance of kernel approximation 
methods to DNNs, demonstrating the effectiveness of using linear bottlenecks, 
performing feature selection, and using the new early stopping criteria in 
bringing down the TER. Lastly, we take a deeper look at the dynamics of the 
feature selection process.

\subsection{Tasks, Datasets, and Evaluation Metrics}
\label{sec:exp_intro}
We train both DNNs and kernel-based multinomial logistic regression models, 
as described in \S\ref{sec:background}, to predict context-dependent HMM 
state labels from acoustic feature vectors.  We test these methods on four 
datasets.

Each dataset is partitioned in four: a training set, a heldout set, a development 
set, and a test set.  We use the heldout set to tune the hyperparameters of 
our training procedure (\eg, the learning rate).  We then run decoding on the 
development set, using IBM's Attila speech recognition toolkit 
\citep{attila2010}, to select a small subset of models which perform best in 
terms of TER (\eg, the best kernel model, and the best DNN model, per dataset). 
We tune the acoustic model weight in order to optimize the relative 
contributions of the language model and the acoustic model to the final score 
our system assigns to a given word sequence. Finally, we decode the 
test set using this select group of models (using, for each model, the best 
acoustic model weight on the development set), 
in order to get a fair comparison between the methods we are 
using.  Having a separate development set helps us avoid the risk of 
over-fitting to the test set.

The first two datasets we use are the IARPA Babel Program Cantonese 
(IARPA-babel101-v0.4c) and Bengali (IARPA-babel103b-v0.4b) limited language packs.
Each pack contains training and development sets of approximately 20 hours, and
an approximately 5 hour test set.  We designate about 10\% of the training data as a heldout 
set.  The training, heldout, development, and test sets all contain different 
speakers.  Babel data is challenging because it is two-person conversations 
between people who know each other well (family and friends) recorded over 
telephone channels (in most cases with mobile telephones) from speakers in a 
wide variety of acoustic environments, including moving vehicles and public 
places.  As a result, it contains many natural phenomena such as 
mispronunciations, disfluencies, laughter, rapid speech, background noise, 
and channel variability.  An additional challenge in Babel is that the only 
data available for training language models is the acoustic transcripts, 
which are comparatively small.  

The third dataset is a 50-hour subset of Broadcast News (BN-50), 
which is a well-studied benchmark task in 
the ASR community \citep{kingsbury09,sainath2011making}.
45 hours of audio are used for training, and 5 hours are used as
a heldout set.  For the development set, we use the ``Dev04F'' dataset 
provided by LDC, which consists of 2 hours of broadcast news from various news 
shows.  We use the DARPA EARS RT-03 English Broadcast News Evaluation Set 
\citep{rt03} as our test set, consisting of 72 5-minute conversations. 

The last dataset we use is TIMIT \citep{timit}, which contains recordings of 
630 speakers, of various English dialects, each reciting ten sentences, for a 
total of 5.4 hours of speech. The training set (from which the heldout set is 
then taken) consists of data from 462 speakers each reciting 8 sentences (SI 
and SX sentences).  The development set consists of speech from 50 speakers.  
For evaluation, we use the ``core test set'', which consists of 192 
utterances total from 24 speakers (SA sentences are excluded).  For 
reference, we use the exact same features, labels, and divisions of the 
dataset as \citet{huang14kernel}, which allows direct comparison of our 
results with theirs.

The language models we use are all $n$-gram language models estimated using 
modified Kneser-Ney smoothing, with $n$ values of $2$,$4$,$3$, and $3$ for 
Bengali, Broadcast News, Cantonese, and TIMIT, respectively.  The TIMIT 
language model is a phone-level model. The Bengali and Cantonese language 
models are particularly small (approximately $ \num[group-separator={,}]{60000}$
bigrams and $\num[group-separator={,}]{136000}$ trigrams, respectively), 
trained using only the provided audio transcripts. The Broadcast 
News model is small as well, containing only 3.3 million $4$-grams.

The acoustic features, representing 25 ms acoustic frames with context, are 
real-valued dense vectors.  A 10 ms shift is used between adjacent frames 
(except on TIMIT, where a 5 ms shift is used).  For the Cantonese, Bengali, 
and Broadcast News datasets we use a standard 360-dimensional speaker-adapted 
representation used by IBM \citep{kingsbury13high}. The state labels are 
obtained via forced alignment using a GMM/HMM system.  For the TIMIT 
experiments, we use 40 dimensional feature space maximum likelihood linear 
regression (fMLLR) features \citep{gales1998}, and concatenate the 5 
neighboring frames in either direction, for a total of 11 frames and 440 
features.

The Cantonese and Bengali datasets each have 1000 labels, corresponding to 
quinphone context-dependent HMM states clustered using decision trees.  For 
Broadcast News, there are 5000 such states. The TIMIT dataset has 147 
context-independent labels, corresponding to the beginning, middle, and end of 49 
phonemes.

For all datasets, the number of training points significantly exceeds typical 
machine learning tasks tackled by kernel methods.  In particular, our training 
sets all contain between 2 and 16 millions frames.  Additionally, the large 
number of output classes for our datasets presents a scalability challenge, 
given that the size of the kernel models scales linearly with the number of 
output classes (if no bottleneck is used). Table~\ref{table:data} provides 
details on the sizes of all the datasets, as well as on their number of features and classes.

\begin{table}
\small
\centering
\begin{tabular}{|c|c|c|c|c|c|c|}\hline
Dataset  & Train & Heldout & Dev  & Test & \# Features & \# Classes \\ \hline
Beng.    & 21 hr  (7.7M)  & 2.8 hr (1.0M) & 20 hr   (7.1M) & 5 hr (1.7M)    & 360 & 1000 \\ \hline
BN-50    & 45 hr  (16M)   & 5 hr   (1.8M) & 2 hr    (0.7M) & 2.5 hr (0.9M)  & 360 & 5000 \\ \hline
Cant.    & 21 hr  (7.5M)  & 2.5 hr (0.9M) & 20 hr   (7.2M) & 5 hr (1.8M)    & 360 & 1000 \\ \hline
TIMIT    & 3.2 hr (2.3M)  & 0.3 hr (0.2M) & 0.15 hr (0.1M) & 0.15 hr (0.1M) & 440 & 147  \\ \hline
\end{tabular}
\caption{Dataset details.  We report the size of each dataset partition in terms
of the number of hours of speech, and in terms of the number of acoustic frames
(in parentheses).}
\label{table:data}
\end{table}

We use five metrics to evaluate the acoustic models:
\begin{enumerate}
\item \textbf{Cross-entropy}: Given examples, \mbox{$\{ (\vx_i, y_i), i = 1 \ldots N\}$}, the cross-entropy is defined as 
$$-\frac{1}{N} \sum_{i=1}^N \log p(y_i | \vct{x}_i).$$

\item \textbf{Average Entropy}: The average entropy of a model is defined as
$$-\frac{1}{N}\sum_{i=1}^N \sum_{y=1}^C  p(y|x_i)\log p(y|x_i)$$
If a model has low average entropy, it is generally confident in its 
predictions.

\item \textbf{Entropy Regularized Log Loss (ERLL)}: Defined in Section 
\ref{sec:method_earlystop}.  We use $\beta=1$ unless specified otherwise.

\item \textbf{Classification Error}: The classification error is defined as 
$$\frac{1}{N} \sum_{i=1}^N \mathbbm{1} \left[ y_i \neq \argmax_{y \in {1,2, \ldots, C}} p(y|\vct{x}_i)\right].$$

\item \textbf{Token Error Rate (TER)}: We feed the predictions of the 
acoustic models, which correspond to probability distributions over the 
phonetic states, to the rest of the ASR pipeline and calculate the 
misalignment between the decoder's outputs and the ground-truth 
transcriptions.  For Bengali and BN-50, we measure the error in terms of the 
word error rate (WER), for Cantonese we use the character error rate (CER), 
and for TIMIT we use the phone error rate (PER).  We use the term ``token 
error rate'' (TER) to refer, for each dataset, to its corresponding metric. 
\end{enumerate}

\subsection{Details of Acoustic Model Training}
\label{sec:exp_train_details}
All our kernel models were trained with either the Laplacian, the Gaussian, 
or the Sparse Gaussian (\S\ref{sec:method-sparse-rbf}) kernel.  These kernel 
models typically have 3 hyperparameters: the kernel bandwidth ($\sigma$ for 
the Gaussian kernels, $\lambda$ for the Laplacian kernel; see Table 
\ref{table:GaussLap}), the number of random projections, and the initial learning 
rate of the optimization procedure. As a rule of thumb, good values for the 
kernel bandwidths (specifically, $2\sigma^2$ for the Gaussian kernels, and 
$1/\lambda$ for the Laplacian kernel) range from 0.3-5 times the median of the 
pairwise distances in the data.\footnote{For the Gaussian kernel, we take the median of the
$\ell_2$ \textit{squared} distances between a large number of random pairs of training examples.  
For the Laplacian kernel, we use $\ell_1$ 
distances instead. For the Sparse Gaussian kernel, we use the median $\ell_2$ squared distances
between randomly chosen sub-vectors of size $k$ of random pairs of training points.} 
We try various numbers of random features, ranging from 
$\num[group-separator={,}]{5000}$ to $\num[group-separator={,}]{200000}$.  
Using more random features 
leads to a better approximation of the kernel function, as well as to more 
powerful models, though there are diminishing returns as the number of 
features increases.  The Sparse Gaussian kernel additionally has the 
hyperparameter $k$ which specifies the sparsity of each random projection 
vector $\vomega_i$.  For all experiments, we use $k=5$.

For all DNNs, we tune hyperparameters related to both the architecture and 
the optimization. This includes the number of layers, the number of hidden 
units in each layer, and the learning rate. We perform 1 epoch of layer-wise 
discriminative pre-training \citep{seide11pretrain,kingsbury13high}, and then 
train the entire network jointly using SGD.  We find that 4 hidden layers is 
generally the best setting for our DNNs, so all the DNN results we present in 
this paper use this setting.  Additionally, all our DNNs use the $\tanh$ 
activation function.  We vary the number of hidden units per layer (1000, 2000, 
or 4000).

For both DNN and kernel models, we use stochastic gradient descent (SGD) as 
our optimization algorithm, with a mini-batch size of 250 or 256 samples.  We 
use the heldout set to tune the other hyperparameters (\eg, learning rate).  
We use the learning rate decay scheme described in 
\citep{morgan1990generalization,sainath2013low,sainath2013b}, which monitors performance on 
the heldout set in order to decide when to decay the learning rate.  This 
method divides the learning rate in half at the end of an SGD epoch if the 
heldout cross-entropy doesn't improve by at least $1\%$; additionally, if the 
heldout cross-entropy gets \textit{worse}, it reverts the model back to its 
state at the beginning of the epoch. Instead of using the heldout cross-entropy,
in some of our experiments we use the heldout ERLL in order to
decide when to decay the learning rate.

As mentioned in Section~\ref{background:bottleneck}, one effective way of reducing 
the number of parameters in our models is to impose a low-rank constraint on 
the output parameter matrix; we refer to this as a ``linear bottleneck'' 
\citep{sainath2013low}.  
We use bottlenecks of size $1000$, $250$, $250$, and $100$ for BN-50, Bengali, 
Cantonese, and TIMIT, respectively.  We train models both with and without 
this technique; the only exception is that we are unable to train BN-50 kernel 
models \textit{without} the bottleneck of size $1000$, due to memory 
constraints on our GPUs.

We initialize our DNN parameters uniformly at random in the range 
$[-\frac{\sqrt{6}}{\sqrt{d_{in} + d_{out}}},\frac{\sqrt{6}}{\sqrt{d_{in} + d_{out}}}]$,
as suggested by \citet{glorot2010}; here, $d_{in}$ and $d_{out}$ refer to 
the dimensionality of the input and output of a DNN layer, respectively. For 
our kernel models, we initialize the random projection matrix as discussed in 
Section~\ref{sec:background}, and we initialize the parameter matrix $\Theta$ as the
zero matrix.  When using a linear bottleneck to decompose the parameter matrix, we 
initialize the resulting two matrices randomly, like we do for our DNNs.

For each iteration of random feature selection, we draw a random subsample of 
the training data of size $R = 10^6$ (except when $D \geq 10^5$, in which case we 
use $R = 2{\times}10^6$, to ensure a safe $n$ to $D$ ratio), but ultimately 
we use all $N$ training examples once the random features are selected.  
Thus, each iteration of feature selection has equivalent computational cost 
to a $R/N$ fraction of an SGD epoch (roughly $1/7$ or $2/7$ for $D < 10^5$ 
and $D \geq 10^5$ respectively, on the Babel speech data sets, for example). We 
use $T = 50$ iterations of feature selection, and in iteration $t$, we select 
$s_t = t\cdot(D/T) = 0.02Dt$ random features. Thus, the total computational 
cost we incur for feature selection is equivalent to approximately seven (or 
14) epochs of training on the Babel data sets.  For the Broadcast News dataset,
it corresponds to the cost of approximately 6 full epochs of training 
(when using $R = 2{\times}10^6$).

All our training code is written in MATLAB, leveraging its GPU features.  We 
execute our code on Amazon EC2 machines, with instances of type g2.2xlarge.  
We use StarCluster\footnote{\url{http://star.mit.edu/cluster}} to more easily 
manage our clusters of EC2 machines.

\subsection{Results}
\label{sec:exp_results}
\begin{table}
\centering
\noindent\makebox[\textwidth]{
\begin{tabular}{*{13}{c|}}
\cline{2-13}
& \multicolumn{4}{c|}{Laplacian} & \multicolumn{4}{c|}{Gaussian} & \multicolumn{4}{c|}{Sparse Gaussian} \\ \cline{2-13}
& NT & B & R & BR & NT & B & R & BR & NT & B & R & BR \\ \hline
\multicolumn{1}{|c|}{Beng.} & 74.5 & 72.1 & 74.5 & 71.4 & 72.6 & 72.0 & 72.6 & 71.8 & 73.0 & 71.5 & 73.0 & \textbf{70.9} \\ \hline
\multicolumn{1}{|c|}{+FS} & 72.9 & 71.1 & 72.8 & 70.4 & 74.1 & 71.4 & 74.2 & \textbf{70.3} & 72.9 & 71.2 & 72.8 & 70.7 \\ \hhline{|=|=|=|=|=|=|=|=|=|=|=|=|=|}
\multicolumn{1}{|c|}{BN-50} & N/A & 17.9 & N/A & 17.7 & N/A & 17.3 & N/A & 17.1 & N/A & 17.3 & N/A & \textbf{17.0} \\ \hline
\multicolumn{1}{|c|}{+FS} & N/A & 17.1 & N/A & \textbf{16.7} & N/A & 17.5 & N/A & 17.0 & N/A & 17.1 & N/A & \textbf{16.7} \\ \hhline{|=|=|=|=|=|=|=|=|=|=|=|=|=|}
\multicolumn{1}{|c|}{Cant.} & 69.9 & 68.2 & 69.2 & 67.4 & 70.2 & 67.6 & 70.0 & \textbf{67.1} & 68.6 & 67.5 & 68.1 & \textbf{67.1} \\ \hline
\multicolumn{1}{|c|}{+FS} & 68.4 & 67.5 & 68.5 & \textbf{66.7} & 69.9 & 67.7 & 69.8 & 66.9 & 68.6 & 67.4 & 68.5 & 66.8 \\ \hhline{|=|=|=|=|=|=|=|=|=|=|=|=|=|}
\multicolumn{1}{|c|}{TIMIT} & 20.6 & 19.2 & 20.4 & 18.9 & 19.8 & 18.9 & 19.6 & 18.6 & 19.9 & 18.8 & 19.6 & \textbf{18.4} \\ \hline
\multicolumn{1}{|c|}{+FS} & 19.5 & 18.6 & 19.3 & 18.4 & 19.5 & 18.6 & 19.4 & 18.4 & 19.3 & 18.4 & 19.1 & \textbf{18.2} \\ \hline
\end{tabular}
}
\caption{Kernel TER Results (development set): This table shows TER results 
for our kernel experiments using either the Laplacian, Gaussian, or Sparse 
Gaussian kernels. `NT' specifies that no ``tricks'' were used during training (no 
bottleneck, no feature selection, no special learning rate decay). A `B' 
specifies that a linear bottleneck was used for the parameter matrix; an `R' 
specifies that entropy regularized log loss was used for learning rate decay 
(so `BR' means both were used). `+FS' specifies that feature selection was 
used for the experiments in that row. The best result for each row is in bold.
} 
\label{table:TER-kernel}
\end{table}

\begin{table}
\centering
\noindent\makebox[\textwidth]{
\begin{tabular}{*{13}{c|}}
\cline{2-13}
& \multicolumn{4}{c|}{1000} & \multicolumn{4}{c|}{2000} & \multicolumn{4}{c|}{4000} \\ \cline{2-13}
& NT & B & R & BR & NT & B & R & BR & NT & B & R & BR \\ \hline
\multicolumn{1}{|c|}{Beng.} & 72.3 & 71.6 & 71.7 & 70.9 & 71.5 & 71.1 & 70.7 & 70.3 & 71.1 & 70.6 & 70.5 & \textbf{70.2} \\ \hline
\multicolumn{1}{|c|}{BN-50} & 18.0 & 17.3 & 17.8 & 17.1 & 17.4 & 16.7 & 17.1 & \textbf{16.4} & 16.8 & 16.7 & 16.7 & 16.5 \\ \hline
\multicolumn{1}{|c|}{Cant.} & 68.4 & 68.1 & 67.9 & 67.5 & 67.7 & 67.7 & 67.2 & \textbf{67.1} & 67.7 & \textbf{67.1} & 67.2 & 67.2 \\ \hline
\multicolumn{1}{|c|}{TIMIT} & 19.5 & 19.3 & 19.4 & 19.2 & 19.0 & 18.9 & 19.2 & 19.2 & \textbf{18.6} & \textbf{18.6} & 18.7 & 18.9 \\ \hline
\end{tabular}
}
\caption{DNN TER Results (development set): This table shows TER results for 
DNNs with 1000, 2000, or 4000 hidden units per layer.  `NT' specifies that no 
``tricks'' were used while training the DNN (no bottleneck, no special learning 
rate decay).  A `B' specifies that a linear bottleneck was used for the 
output parameter matrix; an `R' specifies that entropy regularized log loss was used 
for learning rate decay (so `BR' means both were used). The best result for 
each language is in bold.}
\label{table:TER-dnn}
\end{table}
\begin{table}
\centering
\begin{tabular}{c|c|c|c|c|}
\cline{2-5}
& Beng. (D/K) & BN-50 (D/K) & Cant. (D/K) & TIMIT (D/K) \\ \hline
\multicolumn{1}{|c|}{CE}& \textbf{1.243} / 1.256 & \textbf{2.001} / 2.004 & 1.916 / \textbf{1.883} & 1.056 / \textbf{0.9182} \\ \hline
\multicolumn{1}{|c|}{ENT} & \textbf{0.9079} / 1.082 & \textbf{1.274} / 1.457 & \textbf{1.375} / 1.516 & \textbf{0.447} / 0.5756 \\ \hline
\multicolumn{1}{|c|}{ERLL} & \textbf{2.302} / 2.406 & \textbf{3.548} / 3.625 & \textbf{3.459} / 3.493 & 1.671 / \textbf{1.607} \\ \hline
\multicolumn{1}{|c|}{ERR} & \textbf{0.2887} / 0.2936 & \textbf{0.4887} / 0.4931 & 0.4353 / \textbf{0.4287} & 0.324 / \textbf{0.3085} \\ \hline
\multicolumn{1}{|c|}{TER (dev)} & \textbf{70.2} /  70.3 & \textbf{16.4} /  16.7 &  67.1 / \textbf{66.7} &  18.6 / \textbf{18.2} \\ \hline
\multicolumn{1}{|c|}{TER (test)} & \textbf{69.1} / 69.2 & \textbf{11.7} / 11.9 & 63.7 / \textbf{63.2} & 20.5 / \textbf{20.4} \\ \hline
\end{tabular}
\caption{Table comparing the Best DNN (`D') and kernel (`K') results, across 4
datasets and 6 metrics.  The first 4 metrics are on the heldout set, the 
fifth is on the development set, and the last metric is reported on the test 
set.}
\label{table:summary}
\end{table}

\begin{table}
\centering
\begin{tabular}{c|c|c|}
\cline{2-3}
& Test TER (DNN) & Test TER (Kernel) \\ \hline
\multicolumn{1}{|c|}{\citet{huang14kernel}} & 20.5 & 21.3 \\ \hline
\multicolumn{1}{|c|}{This work} & 20.5 & \textbf{20.4} \\ \hline
\end{tabular}
\caption{Table comparing the Best DNN and kernel results from this work to 
those from \citet{huang14kernel}, on the TIMIT test set.}
\label{table:timit-huang}
\end{table}

In this section, we report results from experiments comparing kernel methods 
to deep neural networks (DNNs) on ASR tasks.  We report results on all 4 
datasets, using various combinations of the methods discussed previously.  
For both DNN and kernel methods, we train models with and without linear 
bottlenecks, and with and without using ERLL to 
determine the learning rate decay.  For our kernel methods, we additionally train 
models with and without using feature selection.  We run experiments with all 
three kernels (Laplacian, Gaussian, Sparse Gaussian) and we use 
$\num[group-separator={,}]{100000}$ random features on all datasets expect for 
TIMIT, where we are able to use $\num[group-separator={,}]{200000}$ random 
features (because the output dimensionality is lower).  As mentioned in the 
previous section, for our DNN experiments, we train models with 4 hidden 
layers,\footnote{As mentioned in Section~\ref{sec:exp_train_details}, we find 
that this is generally the best setting.} using the $\tanh$ activation 
function, and using either 1000, 2000, or 4000 hidden units per layer.  We focus on 
comparing the performance of these methods in terms of TER, but we also 
report results for other metrics.  Unless specified otherwise, all TER results 
are on the development set, and all cross-entropy, entropy, classification 
error, and ERLL results are on the heldout set.

In Tables~\ref{table:TER-kernel} and \ref{table:TER-dnn}, we show our TER 
results for our kernel and DNN models, respectively, across all datasets.  
There are many things to notice about these results. Within the kernel 
models, we see that incorporating a linear bottleneck brings large drops in 
TER across the board.\footnote{Recall that we are unable to train BN-50 kernel 
models \textit{without} using a bottleneck because the resulting models would 
not fit on our GPUs.} Performing feature selection generally improves TER as 
well; we see that it improves TER considerably for the Laplacian kernel, and 
modestly for the Sparse Gaussian kernel.  For the Gaussian kernel, it 
typically helps, though there are several instances in which feature 
selection hurts TER (see Section~\ref{sec:exp_effects_featsel} for discussion).
Second, we see that using heldout ERLL to determine 
when to decay the learning rate helps all our kernel models attain lower TER 
values.  Next, we see that without using feature selection, the Sparse 
Gaussian kernel has the best performance across the board.  After we include 
feature selection, it performs very comparably to the Laplacian kernel with 
feature selection. It is interesting to note that without using feature 
selection, the Gaussian kernel is generally better than the Laplacian kernel; 
however, with feature selection, the Laplacian kernel surpasses the Gaussian 
kernel (see Section~\ref{sec:exp_effects_featsel}).  In general, the kernel 
function which performed best, across the majority of settings, was the 
Sparse Gaussian kernel.

For our DNN models, linear bottlenecks almost always lower TER values, though 
in a few cases they have no effect on TER.  Using ERLL
to determine when to decay the learning rate generally helps lower TER 
values for our DNNs, but in a few cases it actually hurts (Cantonese with 4000
hidden units, and TIMIT with 2000 and 4000).  The DNNs with 4000 hidden 
units typically attain the best results, though on a couple of datasets they are 
matched or narrowly beaten by the 2000 models.

In Table~\ref{table:summary}, we compare for each dataset the performance of the
best DNN model with the best kernel model, across 6 metrics.  In terms of heldout 
cross-entropy and classification error, kernels and DNNs performed similarly, with 
kernels outperforming DNNs on Cantonese and TIMIT, while the DNNs beat the 
kernels on Bengali and BN-50.  In terms of the average heldout entropy of the models, 
the DNNs were consistently more confident in their predictions (lower entropy)
than the kernels. Significantly, we observe that the best development set TER results 
for our DNN and kernel models are quite comparable; on Cantonese and TIMIT, the 
kernel models outperform the DNNs by $0.4\%$ absolute, whereas on Bengali and 
BN-50, the DNN does better by $0.1\%$ and $0.3\%$, respectively.

We will now discuss the results on the test sets.  First of all, in order to avoid 
overfitting to the test set, for each dataset we only performed test set evaluations for the 
DNN and kernel models which performed best in terms of the development set 
TER. The final row of Table~\ref{table:summary} thus contains all the test 
results we collected.  As one can see, the relative performance of the DNN 
and kernel models is very similar to the development set results, with the 
DNNs performing better on Bengali and BN-50, and the kernels performing better 
on Cantonese and TIMIT.  For direct comparison, we include in Table
\ref{table:timit-huang} the test results for the best DNN and kernel models from 
\citet{huang14kernel}.  As mentioned in Section~\ref{sec:exp_intro}, we use 
the same features, labels, data set partitions (train/heldout/dev/test), and 
decoding script as Huang \etal, and thus our results are directly 
comparable.  We achieve a $0.9\%$ absolute improvement in TER with our kernel 
model relative to \citet{huang14kernel}; our DNN performs the same as theirs.
Furthermore, while their DNN beat their kernel by $0.8\%$ TER, our kernel beats
our DNN by $0.1\%$ TER.

In Appendix~\ref{sec:appendixB}, we include more detailed tables comparing 
the various models we trained across all the abovementioned metrics. Some 
important things to take note of in those tables are as follows: 
\begin{itemize} 
\item The linear bottleneck typically causes large drops in the average entropy 
of kernel models, while not having as strong or consistent an effect on 
cross-entropy.  For DNNs, the bottleneck typically causes increases in 
cross-entropy, and relatively modest decreases in entropy. 
\item Using ERLL to determine learning rate decay typically causes increases in 
cross-entropy, and decreases in entropy, with the decrease in entropy typically 
being larger than the increase in cross-entropy.  As a result, the ERLL
is typically lower for models that use this method (with the exception of TIMIT 
DNN models). 
\item Feature selection typically results 
in large drops in cross-entropy, especially for Laplacian and Sparse Gaussian 
kernels, while its effect on entropy is quite small.  It thus helps lower 
heldout ERLL across the board, as well as TER in the vast majority of cases.
\end{itemize}

\subsection{Importance of the Number of Random Features}
\label{sec:exp_results_numRFF}
\begin{figure}[t]
	\centering
    \begin{tabular}{cc}
      \includegraphics[width=0.45\textwidth]{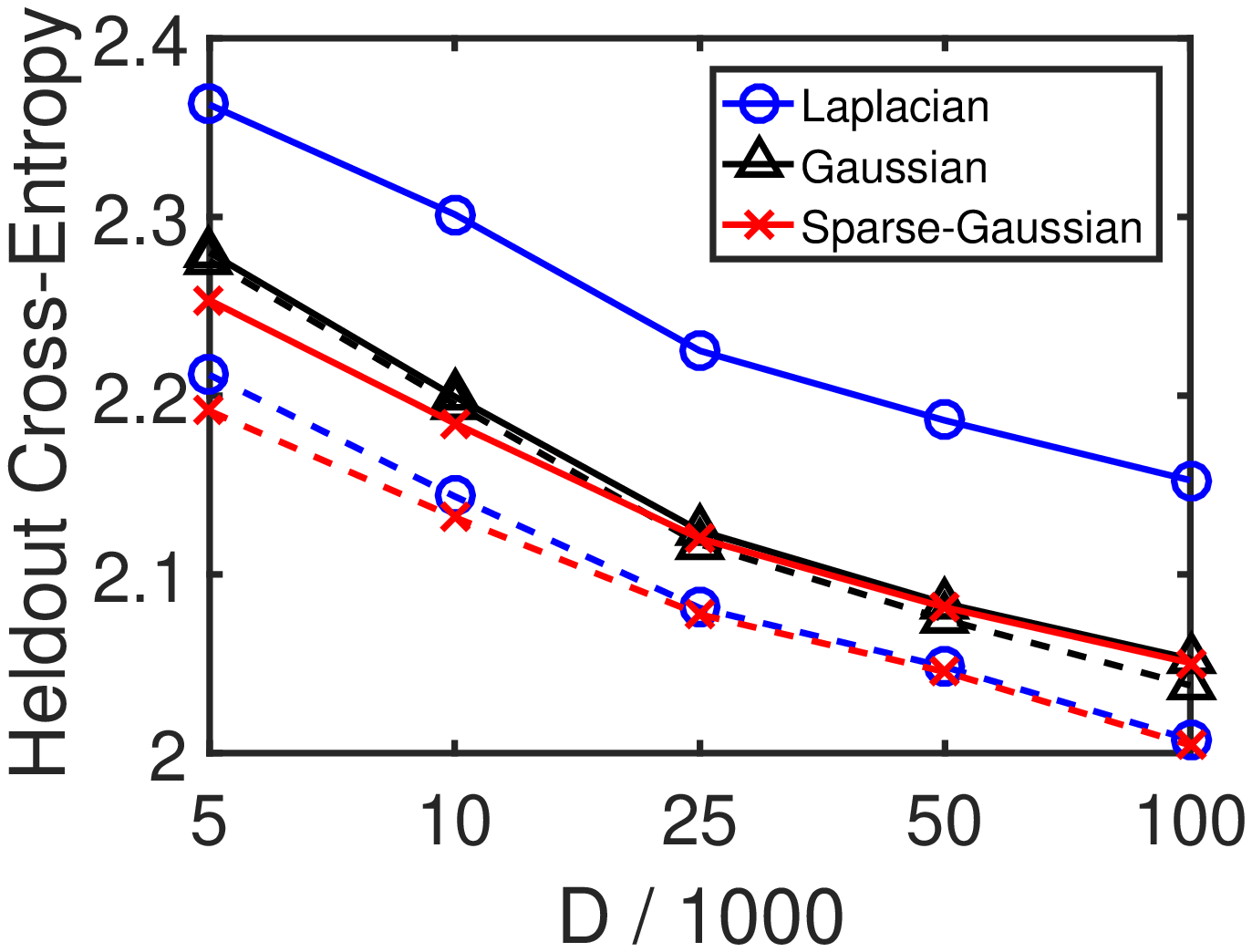} &
			\includegraphics[width=0.45\textwidth]{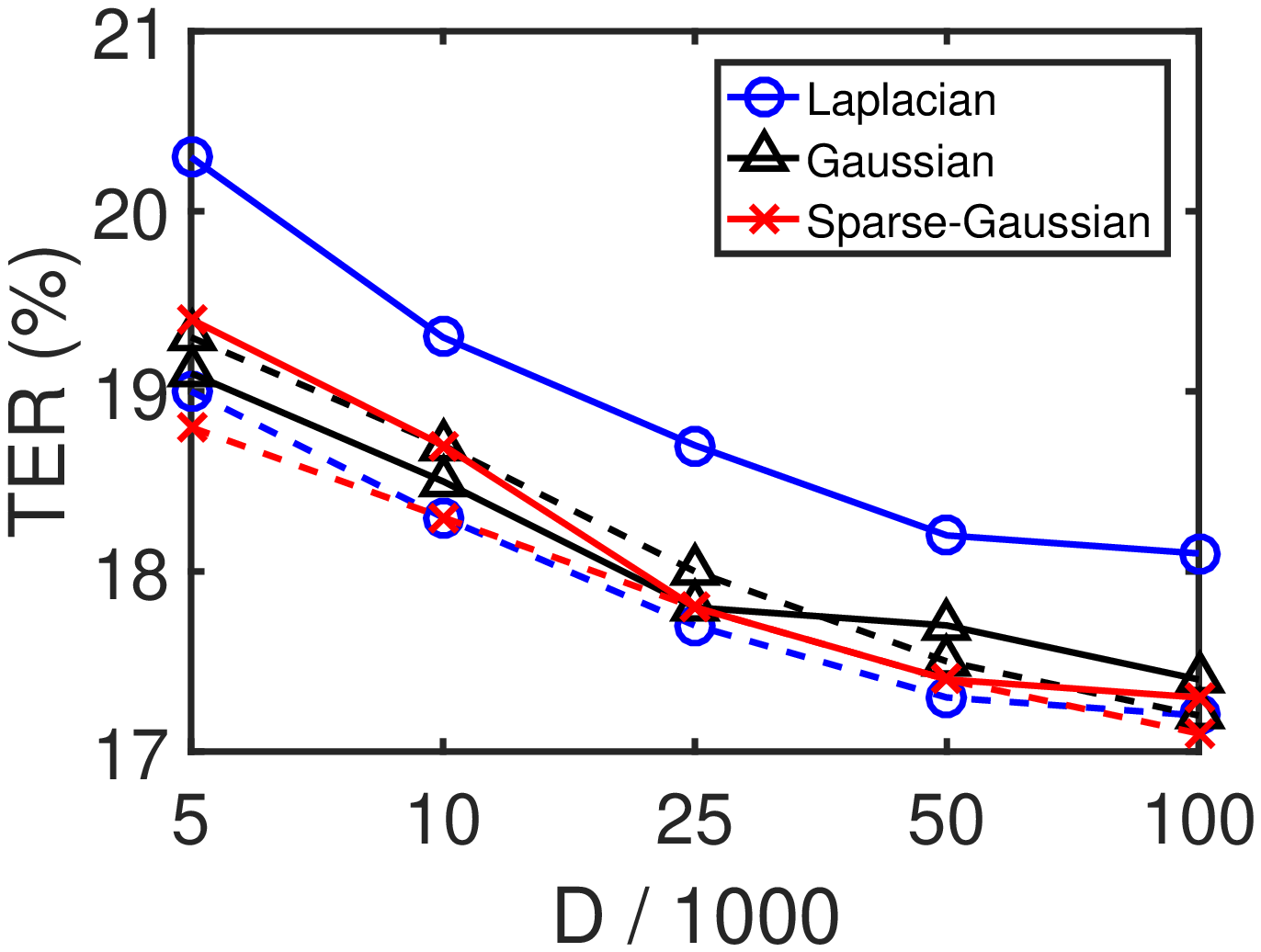}
		\end{tabular}
\caption{Performance of kernel acoustic models on BN-50 dataset, as a
function of the number of random features $D$ used. Results are reported in
terms of heldout cross-entropy as well as development set TER. Dashed lines
signify that feature selection was performed, while solid lines mean it was 
not. The color and shape of the markers indicate the kernel used.}
\label{fig:perf_vs_D} 
\end{figure}

We will now illustrate the importance of the number of random features $D$ on 
the final performance of the model.  For this purpose, we trained a number of 
different models on the BN-50 dataset, using $D \in \{5000,10000,25000,50000,
100000\}$.  We trained models using the 3 different kernels, with and without 
feature selection.  We used a linear bottleneck of size $1000$ for all these 
models, and used heldout cross-entropy to determine the learning rate decay.  
In Figure~\ref{fig:perf_vs_D}, we show how increasing the number of features 
dramatically improves the performance of the learned model, both in terms of 
cross-entropy and TER; there are diminishing returns, however, with very 
small improvements in TER when increasing $D$ from $\num[group-separator={,}]{50000}$
to $\num[group-separator={,}]{100000}$.  Furthermore, the size of the 
gap between the dashed and solid lines (representing experiments with and 
without feature selection, respectively), indicates the importance of feature 
selection in attaining strong performance.  This gap is very large for the 
Laplacian kernel, modest for the Sparse Gaussian kernel, and relatively 
insignificant for the Gaussian kernel.

\subsection{Effects of Random Feature Selection}
\label{sec:exp_effects_featsel}
\begin{figure}
\centering
\includegraphics[width=0.48\textwidth,trim={0 0 0 0},clip]{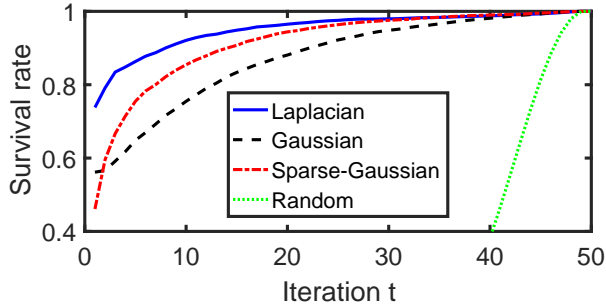}
\caption{Fraction of the $s_t$ features selected in iteration
$t$ that are in the final model (survival rate) for Cantonese dataset.}
\label{fig:survived}
\end{figure}

We now explore the dynamics of the feature selection process. In 
our method, there is no guarantee that a feature selected in one iteration 
will be selected in the next.  In Figure~\ref{fig:survived}, we plot the 
fraction of the $s_t$ features selected in iteration $t$ that actually remain 
in the model after all $T$ iterations. We only show the results for Cantonese 
(models without linear bottleneck, and without using entropy regularized log 
loss for LR decay), as the plots for other datasets are qualitatively 
similar. In nearly all iterations and for all kernels, over half of the 
selected features survive to the final model. For instance, over $90\%$ of 
the Laplacian kernel features selected at iteration $10$ survive the 
remaining $40$ rounds of selection. For comparison, we also plot the expected 
fraction of the $s_t$ features selected in iteration $t$ that would survive 
until the end if the selected features in each iteration were chosen 
uniformly at random from the pool.  Since we use $s_t = Dt/T$, the expected 
fraction at iteration $t$ is $T!/(t! \cdot T^{T-t})$, which is exponentially 
small in $T$ when $t \leq \beta T$ for any fixed $\beta < 1$.\footnote{This can
be shown using Stirling's formula. See \citet{jameson15} for a useful review.}
For example, at $t=10$ the expected survival rate is approximately 
$9\times 10^{-11}$ with $T=50$.

\begin{figure}
\centering
\includegraphics[width=0.48\textwidth]{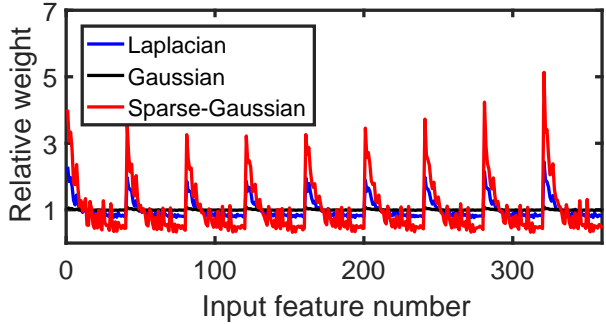}
\caption{The relative weight of each input feature in the random matrix 
$\Theta$, for Cantonese dataset.}
\label{fig:input-features}
\end{figure}

Finally, we consider how the random feature selection process can be regarded as 
selecting non-linear combinations of input features. Consider the final 
matrix of random vectors $\Theta := [\theta^{(1)} | \theta^{(1)} | \dotsb | 
\theta^{(D)}] \in \RR^{d \times D}$ after random feature selection. A coarse 
measure of how much influence an input feature $i \in \{1,2,\dotsc,d\}$ has 
in the final feature map is the relative ``weight'' of the $i$-th row of $
\Theta$.  In Figure~\ref{fig:input-features}, we plot $\sum_{j=1}^D |\Theta_{i,j}|/ Z$ 
for each input feature $i \in \{1,2,\dotsc,d\}$. Here, 
$Z = \frac{1}{d}\sum_{i,j} |\Theta_{i,j}|$ is a normalization term.\footnote{For the Laplacian 
kernel, we discard the largest element in each of the $d$ rows of $\Theta$, 
because there are sometimes outliers which dominate the entire sum for their 
row.} There is a strong periodic effect as a function of the input feature 
number. The reason for this stems from the way the acoustic features are 
generated. Recall that the features are the concatenation of nine 
$40$-dimensional acoustic feature vectors for nine audio frames. An examination of 
the feature pipeline from \citet{kingsbury13high} reveals that these $40$ 
features are ordered by a measure of discriminative quality (via linear 
discriminant analysis). Thus, it is expected that the features with low \mbox
{$(i-1) \bmod 40$} value may be more useful than the others; indeed, this is 
evident in the plot. Note that this effect exists, but is extremely weak, 
with the Gaussian kernel. We believe this is because Gaussian random vectors 
in $\RR^d$ are likely to have \emph{all} their entries be bounded in 
magnitude by $O(\sqrt{\log(d)})$.

\subsection{Other Possible Improvements to DNNs and Kernels}
\label{sec:exp_other_methods}
It is important to mention a few things regarding other ways the performance 
of our DNN and kernel models could be improved, and why they are not 
investigated at length in this work.  For the kernel methods, given that the 
optimization is convex when no bottleneck is used, it would be possible to
get stronger convergence guarantees using the Stochastic Average 
Gradient (SAG) algorithm instead of SGD for training \citep{sag2012}.  In 
fact, in \citet{lu2016} we did this on Cantonese and Bengali, and attained 
strong recognition performance.\footnote{A few more details regarding the
experiments in \citet{lu2016}: we did not use feature selection in that work, 
and we only used ERLL as a model selection criterion (not for learning rate decay).
Additionally, instead of training the large kernel models jointly, we trained 
them in blocks of $\num[group-separator={,}]{25000}$ random features, and then 
combined the models via logit averaging (final models had 
$\num[group-separator={,}]{200000}$ random features).}
Unfortunately, it is challenging to scale this algorithm to larger 
tasks, since it requires storing, for every training example, the previous 
gradient of the loss function at that example.  Because 
$\frac{\partial L(x_i,y_i)}{\partial \vtheta_y} = z(\vx_i)[\mathbb{I}(y = y_i)-p(y|\vx_i)]$, 
and because $z(x_i)$ is fixed, the gradient information can be stored by simply 
storing, for each training example, the vector $\vp_i = [p(1|\vx_i),\ldots,
p(C|\vx_i)]$. However, this still takes $NC$ storage, which is quite expensive 
when there are millions of training examples $N$ and thousands of output 
classes $C$ (320 GB for the Broadcast News dataset, for example). 
Unfortunately, once a bottleneck is introduced, not only is the optimization
problem non-convex, but we must also store the full 
gradients, thus making the memory requirement too large.
As a result, for scalability reasons, as well as for consistency across all 
our experiments, we have used SGD for all our kernel experiments.
Additionally, we did not investigate the use of sequence training techniques 
for our kernel methods, leaving this for future work.

For our DNN models, we have observed that restricted Boltzmann machine (RBM) 
pre-training \citep{hinton06dbn} often improves recognition performance 
\citep{lu2016}.  Additionally, there are various other 
deep architectures (\eg, Convolutional Neural Networks \citep{cnn14}, 
Long Short Term Memory Networks \citep{lstm14}), as well as 
numerous training techniques (\eg, momentum \citep{momentum13}, dropout \citep{dropout14}, 
batch normalization \citep{batch15}), which can further improve the performance of 
neural networks.  Our intention for this paper was to provide a comparison between 
kernel methods and a strong DNN baseline (DNN with $\tanh$ activation, and 
discriminative pre-training), not to provide an exhaustive comparison against 
all possible deep learning architectures and optimization methods.

\section{Conclusion}
\label{sec:conclusion}
In this paper, we explore the performance of kernel methods on large-scale 
ASR tasks, leveraging the kernel approximation technique of \citet{
rahimi07random}.  We propose two new methods (feature selection, 
new early stopping criteria) which lead to large improvements in the performance 
of kernel acoustic models.  We further show that using a linear bottleneck 
\citep{sainath2013low} to decompose the parameter matrix of these kernel models 
leads to significant improvements in TER as well.  We replicate these 
findings on four different datasets, including the Broadcast News (50 hour) 
and TIMIT benchmark tasks.  The linear bottleneck, as well as the learning 
rate decay method, also typically improve the performance of our DNN acoustic 
models.  Using all these methods in conjunction, the kernel methods attain 
comparable TER values to DNNs across our four test sets; 
on Cantonese and TIMIT, the kernel models outperform the DNNs by $0.5\%$ and 
$0.1\%$ absolute, respectively, whereas on Bengali and BN-50, the DNN does better by $0.1\%$ 
and $0.2\%$.

For future work, we are interested in a number of questions: (1) Can we 
develop techniques other than feature selection for ``learning'' the kernel 
function more effectively? (2) Are kernel methods as robust as DNNs to 
different types of input (\eg, log-mel filterbank features)? (3) How much 
does the performance of the kernel models improve using sequence training 
methods? (4) Can sequence kernels be used to improve the recognition 
performance of kernel acoustic models, in a manner analogous to how LSTMs can 
give improvements over DNNs? (5) Can kernel methods compete well with DNNs in 
domains outside of speech recognition? (6) Broadly speaking, what are the 
biggest limitations of kernel methods, and how can they be overcome?

\label{sec:ack}
\acks{This research is supported by the Intelligence Advanced Research Projects 
Activity (IARPA) via 
Department of Defense U.S. Army Research Laboratory (DoD / ARL) contract 
number W911NF-12-C-0012. The U.S. Government is authorized to reproduce and 
distribute reprints for Governmental purposes notwithstanding any copyright 
annotation thereon. Disclaimer: The views and conclusions contained herein 
are those of the authors and should not be interpreted as necessarily 
representing the official policies or endorsements, either expressed or 
implied, of IARPA, DoD/ARL, or the U.S. Government. 

F. S. is grateful to Lawrence K. Saul (UCSD), L\'{e}on Bottou (Facebook),  Alex Smola (Amazon), and Chris J. C. Burges (Microsoft Research) for many fruitful discussions and pointers to relevant work. 

Computation for the work described in this paper was partially supported by the University of Southern California's Center for High-Performance Computing (\url{http://hpc.usc.edu}). 

Additionally, A. B. G. is partially supported by a USC Provost Graduate Fellowship. F. S. is partially supported by a NSF IIS-1065243, 1451412, 1139148 a Google Research Award, an Alfred. P. Sloan Research Fellowship, an ARO YIP Award (W911NF-12-1-0241) and ARO Award W911NF-15-1-0484.  A.B. is partially supported by a grant from CPER Nord-Pas de Calais/FEDER DATA Advanced data science and technologies 2015-2020.
}

\appendix
\section*{Appendices}
\newcommand\numberthis{\addtocounter{equation}{1}\tag{\theequation}}

\section{Derivation of Functional Form for Random Fourier Features}
\label{sec:appendixA}

In this appendix, we will prove that for a properly-scaled (\ie, $Z=1$) positive-definite shift-invariant kernel $k$,
\begin{equation}
k(\vx,\vy) = \expect{\vomega,b}{\sqrt{2}\cos(\vomega\T\vx + b)\cdot \sqrt{2}\cos(\vomega\T\vy + b)},
\label{eq:AppGoal}
\end{equation}
where $\vomega$ is drawn from $p(\vomega)$, the inverse Fourier transform of
$k$, and $b$ is drawn uniformly from $[0,2\pi]$.  We begin this proof using 
Equation \eqref{eq:RFF1} from Section \ref{subsec:RFF}:

\begin{align*}
\hspace{-15mm} k(\vx,\vy)
&= \int_{R^d} p(\vomega) e^{j\vomega\T(\vx-\vy)}\, d\vomega \\
&= \expect{\vomega}{e^{j\vomega\T\vx}e^{-j\vomega\T\vy}}\\
&= \expect{\vomega}{\big(\cos(\vomega\T\vx) + j\sin(\vomega\T\vx)\big)\big(\cos(\vomega\T\vy) - j\sin(\vomega\T\vy)\big)} \\
&= \expect{\vomega}{\cos(\vomega\T\vx)\cos(\vomega\T\vy) + \sin(\vomega\T\vx)\sin(\vomega\T\vy)} \\
&+ j\cdot \expect{\vomega}{\sin(\vomega\T\vx)\cos(\vomega\T\vy) - \sin(\vomega\T\vy)\cos(\vomega\T\vx)} \\
&= \expect{\vomega}{\cos(\vomega\T\vx)\cos(\vomega\T\vy) + \sin(\vomega\T\vx)\sin(\vomega\T\vy)} \numberthis \label{ap1}
\end{align*}

Note the Equation \eqref{ap1} is true because we know that $k(\vx,\vy)$ is a 
real-valued function, and thus the imaginary part of the expectation must 
disappear.  We now show that the right-hand side of Equation \eqref{eq:AppGoal}
is equal to this same expression:

\begin{align*}
&\hspace{-15mm}\expect{\vomega,b}{\sqrt{2}\cos(\vomega\T\vx + b)\cdot \sqrt{2}\cos(\vomega\T\vy + b)} \\
&= 2\cdot\ProbOpr{E}_{\vomega,b}\bigg[\bigg(\cos(\vomega\T\vx)\cos(b)-\sin(\vomega\T\vx)\sin(b)\bigg)\cdot \\
& \hspace{20mm} \bigg(\cos(\vomega\T\vy)\cos(b)-\sin(\vomega\T\vy)\sin(b)\bigg)\bigg] \numberthis \label{ap2}\\
&= 2\cdot\ProbOpr{E}_{\vomega,b}\bigg[
\cos(\vomega\T\vx) \cos(\vomega\T\vy)\cos^2(b) \\
&\hspace{15mm} -\cos(\vomega\T\vx)\sin(\vomega\T\vy)\cos(b)\sin(b) \\
&\hspace{15mm} -\sin(\vomega\T\vx)\cos(\vomega\T\vy)\cos(b)\sin(b) \\
&\hspace{15mm} +\sin(\vomega\T\vx)\sin(\vomega\T\vy)\sin^2(b)\bigg] \\
&= 2\cdot\expect{\vomega}{\frac{1}{2}\cos(\vomega\T\vx)\cos(\vomega\T\vy) + \frac{1}{2}\sin(\vomega\T\vx)\sin(\vomega\T\vy)} \numberthis \label{ap3}\\
&= \expect{\vomega}{\cos(\vomega\T\vx)\cos(\vomega\T\vy) + \sin(\vomega\T\vx)\sin(\vomega\T\vy)} \\
&= k(\vx,\vy)
\end{align*}

Equation \eqref{ap2} is true by the cosine sum of angles formula, and 
Equation \eqref{ap3} is true because 
$\expect{b}{\cos^2(b)}=\expect{b}{\sin^2(b)} = \int_0^{2\pi}{\frac{1}{2\pi}\sin^2(b)} = \frac{1}{2}$, 
and because $\expect{b}{\sin(b)\cos(b)} = 0$.  This concludes the proof. \QED

\FloatBarrier
\section{Detailed Results}

\label{sec:appendixB}
In this appendix, we include tables comparing the models we trained in terms 
of 4 different metrics (CE, ENT, ERR, and ERLL).  The notation is the same as 
in Tables~\ref{table:TER-dnn} and \ref{table:TER-kernel}.  For both DNN and 
kernel models, `NT' specifies that no ``tricks'' were used during training (no 
bottleneck, no feature selection, no special learning rate decay).  A `B' 
specifies that a linear bottleneck was used for the output parameter matrix, 
while an `R' specifies that entropy regularized log loss was used for 
learning rate decay (so `BR' means both were used).  For kernel models, `+FS' 
specifies that feature selection was performed for the corresponding row.  
The best result for each metric and language is in bold.

\begin{table}
\centering
\noindent\makebox[\textwidth]{
\begin{tabular}{*{13}{c|}}
\cline{2-13}
& \multicolumn{4}{c|}{1000} & \multicolumn{4}{c|}{2000} & \multicolumn{4}{c|}{4000} \\ \cline{2-13}
& NT & B & R & BR & NT & B & R & BR & NT & B & R & BR \\ \hline
\multicolumn{1}{|c|}{Beng.} & 1.25 & 1.26 & 1.24 & 1.27 & \textbf{1.24} & 1.26 & 1.26 & 1.32 & 1.24 & 1.25 & 1.30 & 1.39 \\ \hline
\multicolumn{1}{|c|}{BN-50} & 2.05 & 2.05 & 2.04 & 2.08 & 2.01 & 2.04 & 2.05 & 2.22 & \textbf{2.00} & 2.03 & 2.09 & 2.27 \\ \hline
\multicolumn{1}{|c|}{Cant.} & 1.92 & 1.96 & 1.92 & 1.98 & 1.93 & 1.94 & 1.97 & 2.06 & \textbf{1.92} & 1.97 & 2.03 & 2.10 \\ \hline
\multicolumn{1}{|c|}{TIMIT} & \textbf{1.06} & 1.08 & 1.20 & 1.28 & 1.08 & 1.09 & 1.25 & 1.31 & 1.10 & 1.11 & 1.25 & 1.33 \\ \hline
\end{tabular}
}
\caption{DNN: Metric CE}
\end{table}

\begin{table}
\centering
\noindent\makebox[\textwidth]{
\begin{tabular}{*{13}{c|}}
\cline{2-13}
& \multicolumn{4}{c|}{Laplacian} & \multicolumn{4}{c|}{Gaussian} & \multicolumn{4}{c|}{Sparse Gaussian} \\ \cline{2-13}
& NT & B & R & BR & NT & B & R & BR & NT & B & R & BR \\ \hline
\multicolumn{1}{|c|}{Beng.}  & 1.34 & 1.32 & 1.35 & 1.39 & 1.35 & 1.33 & 1.36 & 1.34 & 1.31 & \textbf{1.29} & 1.34 & 1.33 \\ \hline
\multicolumn{1}{|c|}{+FS}  & 1.28 & \textbf{1.26} & 1.29 & 1.27 & 1.35 & 1.31 & 1.36 & 1.35 & 1.28 & 1.26 & 1.31 & 1.27 \\ \hhline{|=|=|=|=|=|=|=|=|=|=|=|=|=|}
\multicolumn{1}{|c|}{BN-50}  & N/A & 2.15 & N/A & 2.43 & N/A & 2.05 & N/A & 2.16 & N/A & \textbf{2.05} & N/A & 2.19 \\ \hline
\multicolumn{1}{|c|}{+FS}  & N/A & 2.01 & N/A & 2.07 & N/A & 2.04 & N/A & 2.13 & N/A & \textbf{2.00} & N/A & 2.06 \\ \hhline{|=|=|=|=|=|=|=|=|=|=|=|=|=|}
\multicolumn{1}{|c|}{Cant.}  & \textbf{1.93} & 1.95 & 1.95 & 2.04 & 1.99 & 1.98 & 2.00 & 2.04 & 1.93 & 1.94 & 1.95 & 2.00 \\ \hline
\multicolumn{1}{|c|}{+FS}  & \textbf{1.88} & 1.90 & 1.89 & 1.95 & 1.97 & 1.97 & 1.98 & 2.03 & 1.90 & 1.91 & 1.91 & 1.96 \\ \hhline{|=|=|=|=|=|=|=|=|=|=|=|=|=|}
\multicolumn{1}{|c|}{TIMIT}  & 0.97 & 0.99 & 0.97 & 1.07 & 0.94 & 0.96 & 0.94 & 1.02 & 0.94 & 0.95 & \textbf{0.94} & 1.03 \\ \hline
\multicolumn{1}{|c|}{+FS}  & 0.92 & 0.95 & 0.92 & 1.03 & 0.93 & 0.96 & 0.93 & 1.02 & \textbf{0.92} & 0.96 & 0.92 & 1.03 \\ \hline
\end{tabular}
}
\caption{Kernel: Metric CE}
\end{table}

\begin{table}
\centering
\noindent\makebox[\textwidth]{
\begin{tabular}{*{13}{c|}}
\cline{2-13}
& \multicolumn{4}{c|}{1000} & \multicolumn{4}{c|}{2000} & \multicolumn{4}{c|}{4000} \\ \cline{2-13}
& NT & B & R & BR & NT & B & R & BR & NT & B & R & BR \\ \hline
\multicolumn{1}{|c|}{Beng.} & 1.23 & 1.17 & 1.18 & 1.09 & 1.18 & 1.13 & 1.09 & 0.99 & 1.14 & 1.11 & 1.02 & \textbf{0.91} \\ \hline
\multicolumn{1}{|c|}{BN-50} & 1.95 & 1.77 & 1.90 & 1.68 & 1.76 & 1.68 & 1.65 & 1.40 & 1.65 & 1.60 & 1.48 & \textbf{1.27} \\ \hline
\multicolumn{1}{|c|}{Cant.} & 1.71 & 1.67 & 1.67 & 1.57 & 1.66 & 1.64 & 1.55 & 1.42 & 1.63 & 1.55 & 1.43 & \textbf{1.38} \\ \hline
\multicolumn{1}{|c|}{TIMIT} & 0.72 & 0.70 & 0.58 & 0.53 & 0.63 & 0.63 & 0.50 & 0.48 & 0.57 & 0.57 & 0.48 & \textbf{0.45} \\ \hline
\end{tabular}
}
\caption{DNN: Metric ENT}
\end{table}

\begin{table}
\centering
\noindent\makebox[\textwidth]{
\begin{tabular}{*{13}{c|}}
\cline{2-13}
& \multicolumn{4}{c|}{Laplacian} & \multicolumn{4}{c|}{Gaussian} & \multicolumn{4}{c|}{Sparse Gaussian} \\ \cline{2-13}
& NT & B & R & BR & NT & B & R & BR & NT & B & R & BR \\ \hline
\multicolumn{1}{|c|}{Beng.} & 1.43 & 1.23 & 1.41 & \textbf{1.08} & 1.36 & 1.31 & 1.35 & 1.28 & 1.35 & 1.23 & 1.30 & 1.10 \\ \hline
\multicolumn{1}{|c|}{+FS} & 1.32 & 1.21 & 1.28 & 1.14 & 1.44 & 1.27 & 1.45 & \textbf{1.13} & 1.32 & 1.22 & 1.26 & 1.14 \\ \hhline{|=|=|=|=|=|=|=|=|=|=|=|=|=|}
\multicolumn{1}{|c|}{BN-50} & N/A & 1.89 & N/A & \textbf{1.46} & N/A & 1.83 & N/A & 1.53 & N/A & 1.81 & N/A & 1.48 \\ \hline
\multicolumn{1}{|c|}{+FS} & N/A & 1.81 & N/A & 1.56 & N/A & 1.84 & N/A & \textbf{1.55} & N/A & 1.80 & N/A & 1.57 \\ \hhline{|=|=|=|=|=|=|=|=|=|=|=|=|=|}
\multicolumn{1}{|c|}{Cant.} & 1.84 & 1.67 & 1.76 & \textbf{1.52} & 1.94 & 1.73 & 1.91 & 1.58 & 1.77 & 1.69 & 1.70 & 1.55 \\ \hline
\multicolumn{1}{|c|}{+FS} & 1.75 & 1.66 & 1.73 & 1.54 & 1.91 & 1.72 & 1.87 & 1.57 & 1.75 & 1.68 & 1.72 & \textbf{1.54} \\ \hhline{|=|=|=|=|=|=|=|=|=|=|=|=|=|}
\multicolumn{1}{|c|}{TIMIT} & 0.95 & 0.72 & 0.91 & 0.61 & 0.88 & 0.73 & 0.86 & 0.62 & 0.89 & 0.76 & 0.85 & \textbf{0.61} \\ \hline
\multicolumn{1}{|c|}{+FS} & 0.86 & 0.70 & 0.82 & 0.58 & 0.86 & 0.70 & 0.83 & 0.61 & 0.84 & 0.69 & 0.82 & \textbf{0.58} \\ \hline
\end{tabular}
}
\caption{Kernel: Metric ENT}
\end{table}

\begin{table}
\centering
\noindent\makebox[\textwidth]{
\begin{tabular}{*{13}{c|}}
\cline{2-13}
& \multicolumn{4}{c|}{1000} & \multicolumn{4}{c|}{2000} & \multicolumn{4}{c|}{4000} \\ \cline{2-13}
& NT & B & R & BR & NT & B & R & BR & NT & B & R & BR \\ \hline
\multicolumn{1}{|c|}{Beng.} & 2.48 & 2.43 & 2.43 & 2.37 & 2.42 & 2.39 & 2.35 & 2.31 & 2.39 & 2.37 & 2.32 & \textbf{2.30} \\ \hline
\multicolumn{1}{|c|}{BN-50} & 3.99 & 3.82 & 3.94 & 3.76 & 3.77 & 3.72 & 3.70 & 3.63 & 3.65 & 3.63 & 3.58 & \textbf{3.55} \\ \hline
\multicolumn{1}{|c|}{Cant.} & 3.63 & 3.63 & 3.58 & 3.56 & 3.59 & 3.58 & 3.51 & 3.48 & 3.55 & 3.52 & \textbf{3.46} & 3.47 \\ \hline
\multicolumn{1}{|c|}{TIMIT} & 1.77 & 1.77 & 1.77 & 1.81 & 1.71 & 1.72 & 1.76 & 1.79 & \textbf{1.67} & 1.68 & 1.73 & 1.78 \\ \hline
\end{tabular}
}
\caption{DNN: Metric ERLL}
\end{table}

\begin{table}
\centering
\noindent\makebox[\textwidth]{
\begin{tabular}{*{13}{c|}}
\cline{2-13}
& \multicolumn{4}{c|}{Laplacian} & \multicolumn{4}{c|}{Gaussian} & \multicolumn{4}{c|}{Sparse Gaussian} \\ \cline{2-13}
& NT & B & R & BR & NT & B & R & BR & NT & B & R & BR \\ \hline
\multicolumn{1}{|c|}{Beng.} & 2.77 & 2.55 & 2.76 & 2.47 & 2.71 & 2.65 & 2.71 & 2.62 & 2.67 & 2.52 & 2.64 & \textbf{2.44} \\ \hline
\multicolumn{1}{|c|}{+FS} & 2.60 & 2.47 & 2.57 & \textbf{2.41} & 2.79 & 2.58 & 2.80 & 2.48 & 2.60 & 2.48 & 2.57 & 2.41 \\ \hhline{|=|=|=|=|=|=|=|=|=|=|=|=|=|}
\multicolumn{1}{|c|}{BN-50} & N/A & 4.04 & N/A & 3.88 & N/A & 3.88 & N/A & 3.69 & N/A & 3.86 & N/A & \textbf{3.67} \\ \hline
\multicolumn{1}{|c|}{+FS} & N/A & 3.82 & N/A & 3.63 & N/A & 3.88 & N/A & 3.67 & N/A & 3.80 & N/A & \textbf{3.62} \\ \hhline{|=|=|=|=|=|=|=|=|=|=|=|=|=|}
\multicolumn{1}{|c|}{Cant.} & 3.77 & 3.62 & 3.71 & 3.56 & 3.94 & 3.71 & 3.91 & 3.62 & 3.71 & 3.63 & 3.65 & \textbf{3.54} \\ \hline
\multicolumn{1}{|c|}{+FS} & 3.63 & 3.56 & 3.63 & \textbf{3.49} & 3.88 & 3.69 & 3.86 & 3.60 & 3.64 & 3.58 & 3.63 & 3.50 \\ \hhline{|=|=|=|=|=|=|=|=|=|=|=|=|=|}
\multicolumn{1}{|c|}{TIMIT} & 1.92 & 1.71 & 1.87 & 1.68 & 1.82 & 1.70 & 1.80 & 1.65 & 1.83 & 1.71 & 1.79 & \textbf{1.64} \\ \hline
\multicolumn{1}{|c|}{+FS} & 1.78 & 1.65 & 1.74 & 1.61 & 1.79 & 1.67 & 1.76 & 1.64 & 1.76 & 1.64 & 1.74 & \textbf{1.61} \\ \hline
\end{tabular}
}
\caption{Kernel: Metric ERLL}
\end{table}

\begin{table}
\centering
\noindent\makebox[\textwidth]{
\begin{tabular}{*{13}{c|}}
\cline{2-13}
& \multicolumn{4}{c|}{1000} & \multicolumn{4}{c|}{2000} & \multicolumn{4}{c|}{4000} \\ \cline{2-13}
& NT & B & R & BR & NT & B & R & BR & NT & B & R & BR \\ \hline
\multicolumn{1}{|c|}{Beng.} & 0.29 & 0.29 & 0.29 & 0.29 & \textbf{0.29} & 0.29 & 0.29 & 0.30 & 0.29 & 0.29 & 0.29 & 0.30 \\ \hline
\multicolumn{1}{|c|}{BN-50} & 0.50 & 0.50 & 0.50 & 0.50 & 0.49 & 0.50 & 0.50 & 0.51 & \textbf{0.49} & 0.49 & 0.50 & 0.51 \\ \hline
\multicolumn{1}{|c|}{Cant.} & 0.44 & 0.44 & \textbf{0.44} & 0.44 & 0.44 & 0.44 & 0.44 & 0.44 & 0.44 & 0.44 & 0.44 & 0.44 \\ \hline
\multicolumn{1}{|c|}{TIMIT} & 0.33 & 0.33 & 0.34 & 0.34 & 0.33 & 0.33 & 0.34 & 0.34 & 0.33 & \textbf{0.32} & 0.33 & 0.33 \\ \hline
\end{tabular}
}
\caption{DNN: Metric ERR}
\end{table}

\begin{table}
\centering
\noindent\makebox[\textwidth]{
\begin{tabular}{*{13}{c|}}
\cline{2-13}
& \multicolumn{4}{c|}{Laplacian} & \multicolumn{4}{c|}{Gaussian} & \multicolumn{4}{c|}{Sparse Gaussian} \\ \cline{2-13}
& NT & B & R & BR & NT & B & R & BR & NT & B & R & BR \\ \hline
\multicolumn{1}{|c|}{Beng.} & 0.30 & 0.30 & 0.31 & 0.31 & 0.31 & 0.31 & 0.31 & 0.31 & 0.30 & \textbf{0.30} & 0.30 & 0.30 \\ \hline
\multicolumn{1}{|c|}{+FS} & 0.29 & \textbf{0.29} & 0.30 & 0.29 & 0.31 & 0.31 & 0.31 & 0.31 & 0.30 & 0.29 & 0.30 & 0.30 \\ \hhline{|=|=|=|=|=|=|=|=|=|=|=|=|=|}
\multicolumn{1}{|c|}{BN-50} & N/A & 0.52 & N/A & 0.54 & N/A & \textbf{0.50} & N/A & 0.51 & N/A & 0.50 & N/A & 0.51 \\ \hline
\multicolumn{1}{|c|}{+FS} & N/A & 0.49 & N/A & 0.50 & N/A & 0.50 & N/A & 0.50 & N/A & \textbf{0.49} & N/A & 0.49 \\ \hhline{|=|=|=|=|=|=|=|=|=|=|=|=|=|}
\multicolumn{1}{|c|}{Cant.} & \textbf{0.43} & 0.44 & 0.44 & 0.44 & 0.45 & 0.45 & 0.45 & 0.45 & 0.43 & 0.44 & 0.44 & 0.44 \\ \hline
\multicolumn{1}{|c|}{+FS} & \textbf{0.43} & 0.43 & 0.43 & 0.44 & 0.44 & 0.44 & 0.44 & 0.45 & 0.43 & 0.44 & 0.43 & 0.44 \\ \hhline{|=|=|=|=|=|=|=|=|=|=|=|=|=|}
\multicolumn{1}{|c|}{TIMIT} & 0.32 & 0.32 & 0.32 & 0.33 & 0.31 & 0.32 & 0.31 & 0.32 & 0.31 & 0.31 & \textbf{0.31} & 0.32 \\ \hline
\multicolumn{1}{|c|}{+FS} & 0.31 & 0.31 & 0.31 & 0.31 & 0.31 & 0.31 & 0.31 & 0.32 & 0.31 & 0.31 & \textbf{0.31} & 0.31 \\ \hline
\end{tabular}
}
\caption{Kernel: Metric ERR}
\end{table}

\FloatBarrier

\vskip 0.2in
\bibliographystyle{plainnat}
\bibliography{ref}

\end{document}